\documentclass[10pt, logo, twocolumn, copyright]{nvidiatechreport}

\usepackage{mdframed}
\usepackage[utf8]{inputenc} 
\usepackage[T1]{fontenc}    

\usepackage{amsfonts}       
\usepackage{nicefrac}       
\usepackage{microtype}      
\usepackage[dvipsnames]{xcolor}         
\usepackage{multirow}
\usepackage{multicol}
\usepackage{graphicx}
\usepackage[numbers]{natbib}
\usepackage{tabto}
\usepackage{xspace}
\usepackage{amsmath}
\usepackage{adjustbox}
\usepackage{enumitem}
\usepackage{wrapfig}
\usepackage{dblfloatfix}
\usepackage{tcolorbox}
\usepackage{listings}
\usepackage{footmisc}

\usepackage{float}

\usepackage{natbib}

\usepackage{hyperref}
\usepackage{url}
\usepackage{booktabs}
\usepackage{multirow}
\usepackage{enumitem}
\usepackage{adjustbox}
\usepackage{tabularx}
\usepackage{arydshln}
\usepackage{wrapfig}

\usepackage{multirow}
\usepackage{colortbl}
\usepackage{amsmath}
\usepackage{makecell}
\usepackage{hhline}
\usepackage{array}
\usepackage{diagbox}
\usepackage{graphicx}
\usepackage{adjustbox}

\usepackage{fp} 

\usepackage{xcolor}
\definecolor{mygreen}{rgb}{0.7, 1.0, 0.7}

\definecolor{myblue}{rgb}{0.7, 0.8, 1.0}

\usepackage{pifont}
\newcommand\mynuma[1]{\ifcase#1 \or \ding{172}\or \ding{173}\or
  \ding{174}\or \ding{175}\or \ding{176}\or \ding{177}%
  \or \ding{178}\or \ding{179}\or \ding{180}\or \ding{181}\else *\fi\relax}

\newcommand\mynumb[1]{\ifcase#1 \or \ding{182}\or \ding{183}\or
  \ding{184}\or \ding{185}\or \ding{186}\or \ding{187}%
  \or \ding{188}\or \ding{189}\or \ding{190}\or \ding{191}\else *\fi\relax}

\newcounter{mycounter} 

\newcommand{\findingbox}[1]{
    \stepcounter{mycounter} 
    \begin{tcolorbox}[colframe=black,
                      arc=1pt,
                      boxsep=-2pt,
                      ]
        \noindent{\textbf{\textit{Finding \themycounter.}}} #1
    \end{tcolorbox}
}

\newcommand{\B}[1]{\textbf{#1}}
\newcommand{\cov}[1]{\mathbf{\Sigma}\left[\mathbf{#1}\right]}
\DeclareMathOperator{\var}{\textup{Var}}
\DeclareMathOperator{\mse}{\textup{MSE}}

\def\paperID{2805} 
\def\confName{CVPR}
\def\confYear{2025}

\title{RADIOv2.5: Improved Baselines for Agglomerative Vision Foundation Models}

\author{Greg Heinrich$^*$ \qquad Mike Ranzinger$^*$ \qquad Hongxu  (Danny) Yin \qquad Yao Lu \qquad Jan Kautz \qquad
Andrew Tao \qquad Bryan Catanzaro \qquad Pavlo Molchanov
}

\begin{document}

\begin{abstract}
Agglomerative models have recently emerged as a powerful approach to training vision foundation models, leveraging multi-teacher distillation from existing models such as CLIP, DINO, and SAM.
This strategy enables the efficient creation of robust models, combining the strengths of individual teachers while significantly reducing computational and resource demands. In this paper, we thoroughly analyze state-of-the-art agglomerative models, identifying critical challenges including resolution mode shifts, teacher imbalance, idiosyncratic teacher artifacts, and an excessive number of output tokens. To address these issues, we propose several novel solutions: multi-resolution training, mosaic augmentation, and improved balancing of teacher loss functions. Specifically, in the context of Vision Language Models, we introduce a token compression technique to maintain high-resolution information within a fixed token count. We release our top-performing variants at multiple scales (-B, -L, -H, and -g), along with inference code and pretrained weights. 

\vspace{5pt}

\textbf{Links:} \hspace{2pt} \href{https://github.com/NVlabs/RADIO}{Code} (on GitHub) | \href{https://huggingface.co/collections/nvidia/radio-669f77f1dd6b153f007dd1c6}{Models} (on Hugging Face) 

\vspace{10pt}

\end{abstract} 
\maketitle
\renewcommand{\thefootnote}{\fnsymbol{footnote}}
\footnotetext[1]{Equal Contribution}
\renewcommand{\thefootnote}{\arabic{footnote}}
\section{Introduction}
\label{sec:intro}

\begin{figure*}[ht]
\begin{center}
  \centering
  \resizebox{\linewidth}{!}{
  \includegraphics[width=0.65\linewidth]{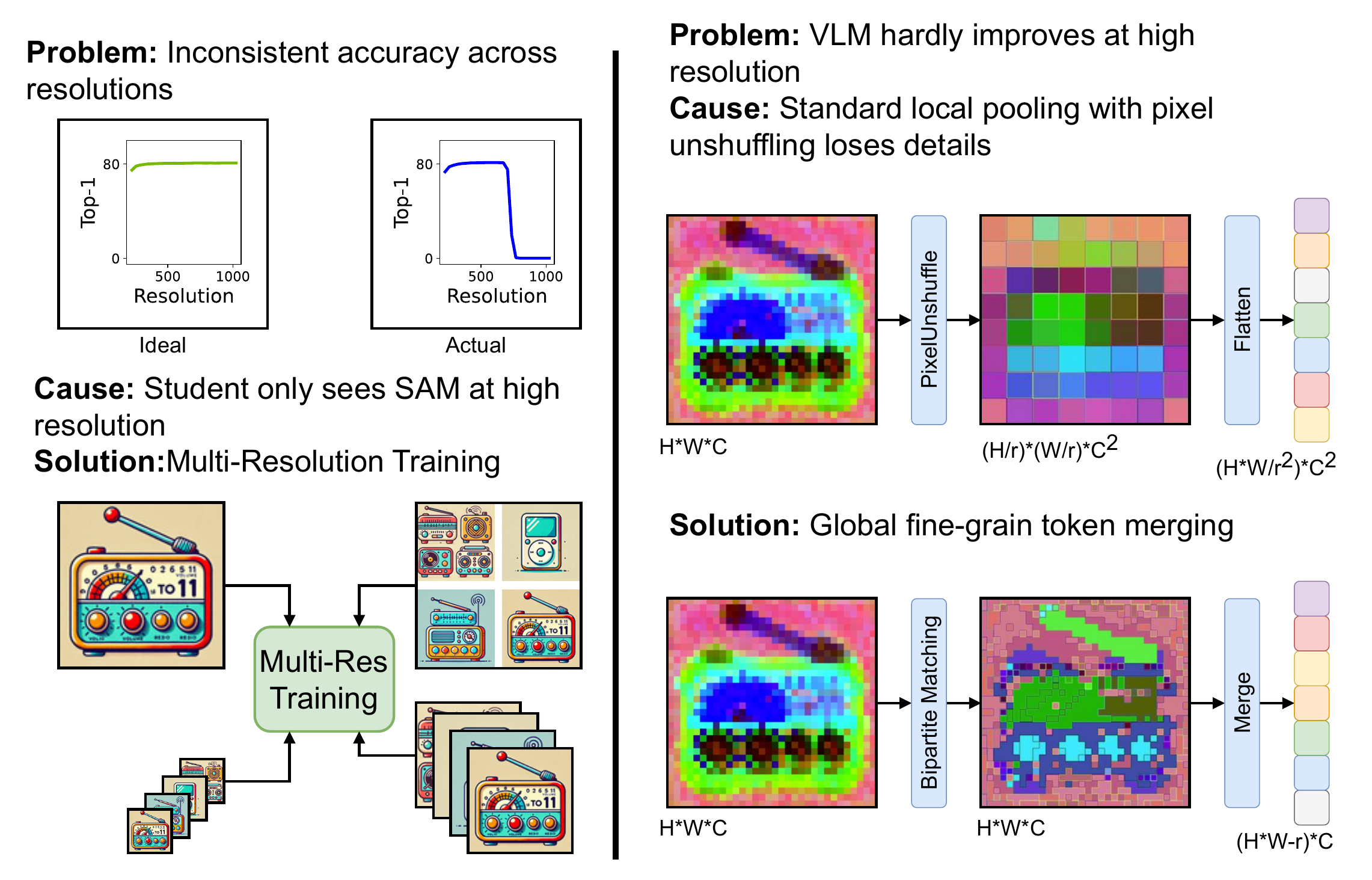}
  \includegraphics[width=0.33\linewidth]{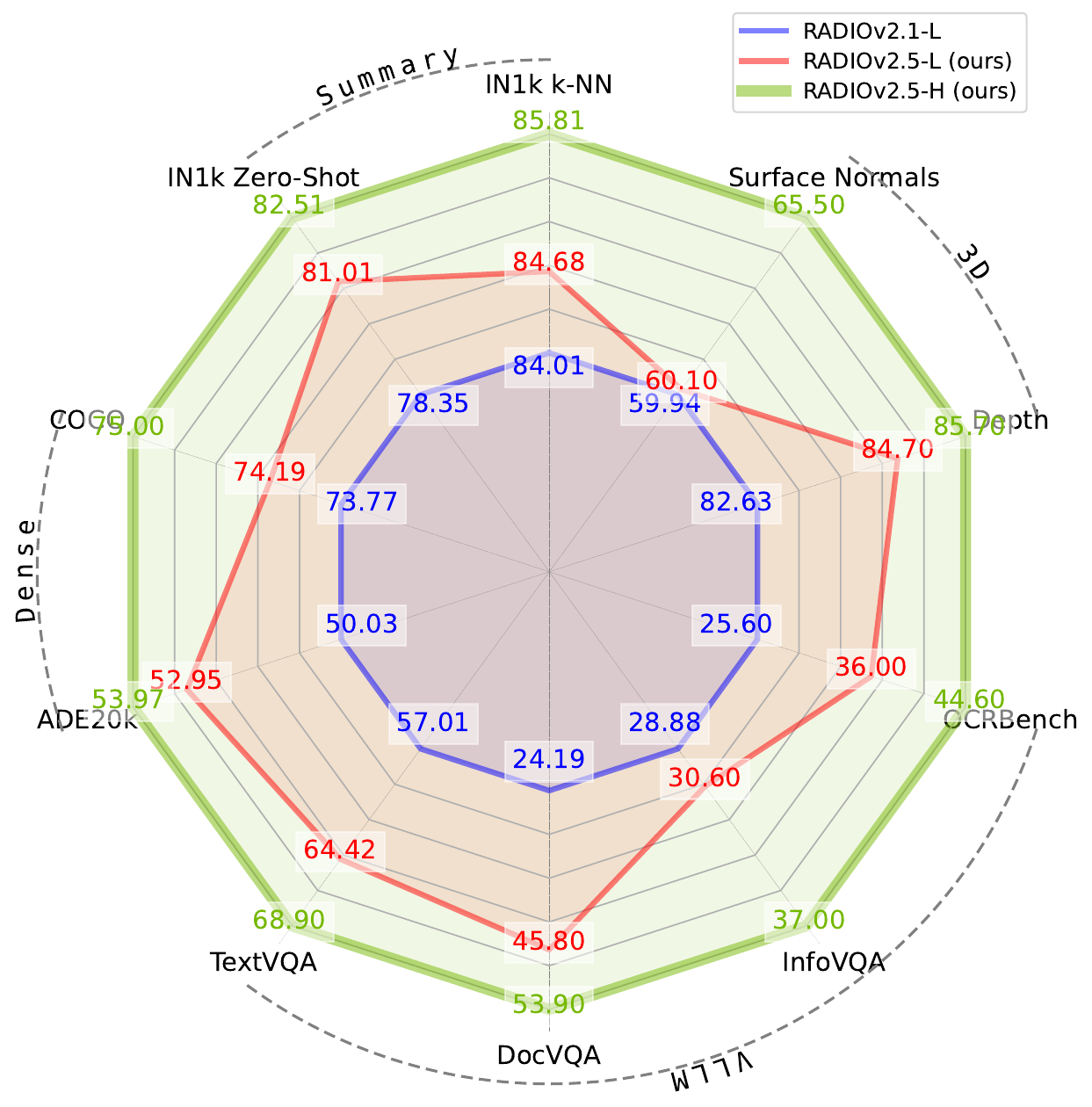}
  }
  \captionof{figure}{Overview of the main contributions: From left to right, we introduce a multi-resolution training framework that enables our student model to maintain accuracy across all resolutions; we propose using token merging to better retain fine-grained details while merging uninformative tokens; our \B{RADIOv2.5} models improve upon the RADIOv2.1 baseline across all benchmarks, with significant gains on VLM tasks.}
  \label{fig:main-contributions}
\end{center}
\end{figure*}

The rise of specialized Vision Foundation Models (VFMs) has created a need for methods to consolidate knowledge from multiple models into a unified model.
SAM-CLIP \cite{wang2024samclipmergingvisionfoundation} addresses this challenge by combining SAM \cite{kirillov2023sam} and CLIP \cite{radford2021learningtransferablevisualmodels} to integrate capabilities from both. In another approach, AM-RADIO \cite{ranzinger2024radio} introduces label-free knowledge distillation from multiple teacher models, enabling knowledge transfer without direct supervision. UNIC \cite{sariyildiz2024unicuniversalclassificationmodels} adds intermediate teacher matching projectors and dynamic teacher selection. Theia \cite{shang2024theiadistillingdiversevision} aims to facilitate robot learning by distilling insights from multiple vision teachers. Meanwhile, PHI-S \cite{ranzinger2024phisdistributionbalancinglabelfree} studies the importance of normalizing the distinct teacher distributions to simplify their balancing. Relatedly, Eagle \cite{shi2024eagleexploringdesignspace} leverages a mixture of vision encoders to achieve inference-time knowledge aggregation within the context of Vision-Language Models (VLMs).

Despite these advancements, this growing body of work on knowledge agglomeration still leaves open several critical challenges:

\begin{itemize}

\item Resolution balancing: Teacher models operate at varying resolutions due to different architectures and training goals, creating feature granularity inconsistencies. Effective techniques are needed to balance these resolutions in the student model to capture both fine details and broader abstractions.

\item Teacher distribution balancing: Existing models have different distribution moments and the distillation process should account for this to prevent biased learning. 

\item Generating multi-resolution features for diverse applications: Vision models support various applications requiring different feature resolutions, from image captioning to dense segmentation.
A VFM that flexibly produces features at any resolution could serve multiple tasks, reducing the need for separate models and unlocking new opportunities.
\end{itemize}

In our analysis of existing agglomerative models, we study and propose a fix for the notable ``mode switch'' phenomenon in AM-RADIO, where feature distributions shift significantly based on input resolution (Section \ref{sec:mode_switch}). Specifically, low-resolution inputs yield DINO-like features, while high-resolution inputs produce SAM-like features. We trace this behavior to the student learning from different teachers at different resolutions during training. In section \ref{sec:mulires}, we introduce a solution to stabilize these mode switches, achieving strong resolution robustness and improved quality scaling.

Armed now with a vision encoder that works best at high resolution, we next look toward integrating it into VLMs. In the case of downstream applications with vision-language models, a common pitfall is the number of output tokens/features. By processing images in high resolution, most methods will return more image tokens and will result in quadratic complexity of attention in VLMs. High resolution processing is important, and in order to preserve this information, we propose applying ToMeSD~\cite{bolya2023tokenmergingfaststable} in section \ref{sec:token_compression} to decouple the vision encoder resolution from the number of patches used by the VLM.

\noindent Our main contributions, illustrated in Figure~\ref{fig:main-contributions}, are:
\begin{itemize}
    \item A multi-resolution training strategy that fixes mode switching and allows for fully flexible input resolution.
    \item A comprehensive study of feature selection for a range of downstream tasks.
    \item A new method to compress visual features, enabling integration with language models while preserving essential information. We demonstrate that our proposed vision encoder disproportionately benefits from this.
\end{itemize}

\begin{table*}
\centering\resizebox{\linewidth}{!}{
    \begin{tabular}{l|l|cccccccccccc}
        \toprule
        \multirow{2}{*}{\textbf{Configuration}} & \multirow{2}{*}{\textbf{Goal}} & \multicolumn{2}{c|}{\B{ImageNet1K}} & \multicolumn{1}{c|}{\B{Segmentation}}
                                        & \multicolumn{2}{c|}{\B{Probe3D\cite{elbanani2024probing}}}
                                        & \multicolumn{5}{c|}{\B{Vision-Language (VILA\cite{lin2023vila})}}
                                        & \multicolumn{1}{c}{\B{SAM~\cite{kirillov2023sam}}} \\ 
                                        &  & Zero-shot  & \multicolumn{1}{c|}{k-NN}  &
                                        \multicolumn{1}{c|}{ADE20k}   
                                        & Depth
                                        & \multicolumn{1}{c|}{SurfNormals}
                                        & TextVQA  & ChartQA & DocVQA  & InfoVQA & \multicolumn{1}{c|}{OCRBench}  & COCO \\
        \midrule
        \B{$\mathcal{A}$:} RADIOv2.1-L*  & Baseline & 78.35 & 84.01 & 50.03 & 82.63 & 59.94   &  57.01  & 15.6
                 & 24.19 &  28.88 & 25.60 & 73.77 \\
        \B{$\mathcal{B}$:} $\mathcal{A}$ + multi-res & Eliminate modes & 81.21 & 84.09 & 52.84 & 82.39 & 61.04 & 59.44 & 16.6 & 33.99 &  29.02 & 29.40 & 75.49 \\
        \B{$\mathcal{C}$:} $\mathcal{B}$ - OpenAICLIP + SigLIP & Better VLM & 81.01 & 84.68 & 52.95  & 84.7 & 60.1 &  64.42 & 25.28 &  45.80 & 30.60 & 36.00 & 74.19 \\
        \B{$\mathcal{D}$:} $\mathcal{C}$ + ViT-H & Bigger backbone & \B{82.51} & \B{85.81} & \B{53.97} & \B{85.7} & \B{62.5} & 65.88 & 25.96 & 49.74 & 35.17 & 40.90 & \B{76.14} \\
        \B{$\mathcal{E}$:} $\mathcal{D}$ + Token Merging & Improve VLM & - & - &  - & - & - & \B{69.74} & \B{30.40} & \B{52.33} & \B{36.24} & \B{42.90} & -  \\
        \bottomrule
    \end{tabular}
    }
    
        \caption{Ablation Results. For Probe3D, Depth and Surface Normals metrics are averaged over the buckets defined in the paper. For SAM COCO we use the ``instance all'' bucket. Each incremental change we introduce leads to improved metrics. *We use a ViT-L instead of the ViT-H used in the original AM-RADIO~\cite{ranzinger2024radio} paper.}
    \label{tab:ablation_table}
    \vspace{-4mm}
\end{table*}

\section{Background}

\subsection{Knowledge Agglomeration}

The assumption underlying knowledge agglomeration is that multiple foundation models exist, each capable of extracting diverse and meaningful representations from a wide range of internet-sourced images. Furthermore, it assumes that knowledge from these models can be distilled into a single agglomerative model.

Let \( x \) be the input data. The student's shared backbone produces a feature representation:
\begin{equation}
\mathbf{z} = f(x) = \Big[z_s,\, z_p^{(1)},\, z_p^{(2)},\, \dots,\, z_p^{(N)}\Big],
\label{eq:backbone}
\end{equation}
where:
\begin{itemize}
    \item \( z_s \in \mathbb{R}^{d} \) is the \emph{summary token},
    \item \( z_p^{(i)} \in \mathbb{R}^{d} \) for \( i=1,\dots,N \) are the \emph{patch tokens},
    \item $d$ is the student's embedding dimension.
\end{itemize}

For each teacher \( t \), the student's model includes two adaptor heads:
\begin{enumerate}
    \item \( g_s^{(t)}: \mathbb{R}^{d} \to \mathbb{R}^{d_t} \), which projects the summary token \( z_s \) into the teacher's embedding space.
    \item \( g_p^{(t)}: \mathbb{R}^{d} \to \mathbb{R}^{d_t} \), which projects each patch token \( z_p^{(i)} \).
\end{enumerate}
These adaptor heads are typically implemented as simple multi-layer perceptrons (MLPs) that can adjust the feature dimension as required. For example, one may define:
\begin{equation}
g^{(t)}(z) = \mathrm{MLP}^{(t)}(z) = \sigma\Big(W^{(t)} z + b^{(t)}\Big),
\label{eq:mlp}
\end{equation}
where \( \sigma(\cdot) \) is a non-linear activation function, and \( W^{(t)} \) and \( b^{(t)} \) are learnable parameters.

Thus, for each teacher \( t \), the projected features are:
\begin{align}
\text{Summary Projection:} \quad & \hat{z}_s^{(t)} = g_s^{(t)}(z_s), \label{eq:summary}\\[1ex]
\text{Patch Projection:} \quad & \hat{z}_p^{(t)} = g_p^{(t)}(z_p). \label{eq:patch}
\end{align}

These projected features are then used in a knowledge distillation process that encourages the student to mimic the representations of each teacher, effectively aggregating diverse knowledge into a single agglomerative model.

The training objective is to align the student features with the corresponding teacher features. To achieve this, we define a loss function that computes an aggregate measure of similarity between the teacher and student features. Let $\mathcal{L}_t$ denote the loss for teacher $t$. This loss can be defined as:
\begin{equation}
\mathcal{L}_t = \ell_s\Big(\hat{z}_s^{(t)}, z_{s}^{(t)}\Big) + \sum_{i=1}^{N} \ell_p\Big(\hat{z}_p^{(t,i)}, z_{p}^{(t,i)}\Big),
\label{eq:teacher_loss}
\end{equation}
where:
\begin{itemize}
    \item $\ell_\cdot(\cdot, \cdot)$ is a similarity or distance metric (e.g., mean squared error or cosine similarity loss)
    \item $\ell_s$, the summary loss objective, need not be the same as $\ell_p$, the patch loss objective
    \item $z_{s}^{(t)}$ and $z_{p}^{(t,i)}$ are the summary and patch features extracted from teacher $t$.
\end{itemize}

The overall loss function aggregates the losses from all teachers:
\begin{equation}
\mathcal{L} = \sum_{t} \lambda_t \, \mathcal{L}_t,
\label{eq:total_loss}
\end{equation}
where $\lambda_t$ are weighting factors that balance the contribution of each teacher's loss.

\subsection{Baseline Model}

Following AM-RADIO\cite{ranzinger2024radio}, our baseline model consists of a Vision Transformer (ViT)\cite{dosovitskiy2021imageworth16x16words} backbone, including CPE~\cite{kim2023regionawarepretrainingopenvocabularyobject} for multi-resolution support. We use DFN~CLIP~\cite{fang2023datafilteringnetworks}, OpenAI~CLIP~\cite{radford2021learningtransferablevisualmodels}, DINOv2-g-reg~\cite{darcet2024visiontransformersneedregisters} and SAM-H~\cite{kirillov2023sam} as teachers. For each teacher to distill from, we augment the backbone with an adaptor for the summarization token, and an adaptor for the patch tokens. Our adaptor is a 2-layer MLP with a LayerNorm\cite{ba2016layernormalization} and a GeLU\cite{hendrycks2023gaussianerrorlinearunits} in between. We train the student using teacher features generated by inferring images of the DataComp1B\cite{gadre2023datacompsearchgenerationmultimodal} dataset. For faster iterations, we replicate AM-RADIO using a ViT-L backbone instead of the original ViT-H backbone and refer to it as \B{RADIOv2.1-L} from this point forward. We train the model with a batch size of 1024+128 for 600k iterations. Training is split into two concurrent partitions: one partition trains the student at a resolution of $432^2$ against CLIP and DINOv2 teachers with batch size 1024, while the other partition trains the student at a resolution of $1024^2$ against SAM with a batch size of 128. This baseline is referred to as \B{$\mathcal{A}$} in Table~\ref{tab:ablation_table}.

\subsection{Evaluation Framework}

Evaluating the quality of a foundation model is a daunting task due to the wide spectrum of downstream applications. We employ a rigorous evaluation framework to identify areas for improvement and guide design decisions:

\begin{itemize}
\item
\textbf{Image-level reasoning:} We focus on ImageNet-1k~\cite{russakovsky2015imagenetlargescalevisual} Top-1 classification accuracy using \textit{(i)} zero-shot~\cite{radford2021learningtransferablevisualmodels} classification using a pretrained and frozen language model; and \textit{(ii)} k-NN~\cite{caron2021emergingpropertiesselfsupervisedvision} classification.
\item
\textbf{Pixel-level semantic segmentation:} We linearly probe the frozen backbone features for semantic segmentation on both ADE20k~\cite{zhou2018semanticunderstandingscenesade20k} and Pascal~VOC~\cite{everingham2010voc}.
\item
\textbf{Instance segmentation:} We evaluate our model on the COCO~\cite{kim2023regionawarepretrainingopenvocabularyobject} dataset using the protocol from EfficientViT~\cite{liu2023efficientvitmemoryefficientvision}, and our learned SAM adaptor.
\item
\textbf{3D understanding:} We consider depth estimation, surface normals estimation, multi-view correspondence, and semantic correspondence, following \cite{elbanani2024probing}.
\item
\textbf{Vision-Language reasoning:} We pair our backbone with an LLM to evaluate Vision-Language modeling.
\item Additional dense multi-task probing on Pascal Context \cite{mottaghi14pascalcontext} and NYUDv2 \cite{silberman12nyudv2} following the setup in MLoRE \cite{jiang2024mlore}.
\end{itemize}
 
\section{Challenges}

\subsection{Achieving Multi-Resolution Robustness}\label{sec:mode_switch}

\begin{figure*}[t]
  \centering
  \resizebox{\linewidth}{!}{
  \includegraphics[width=\linewidth]{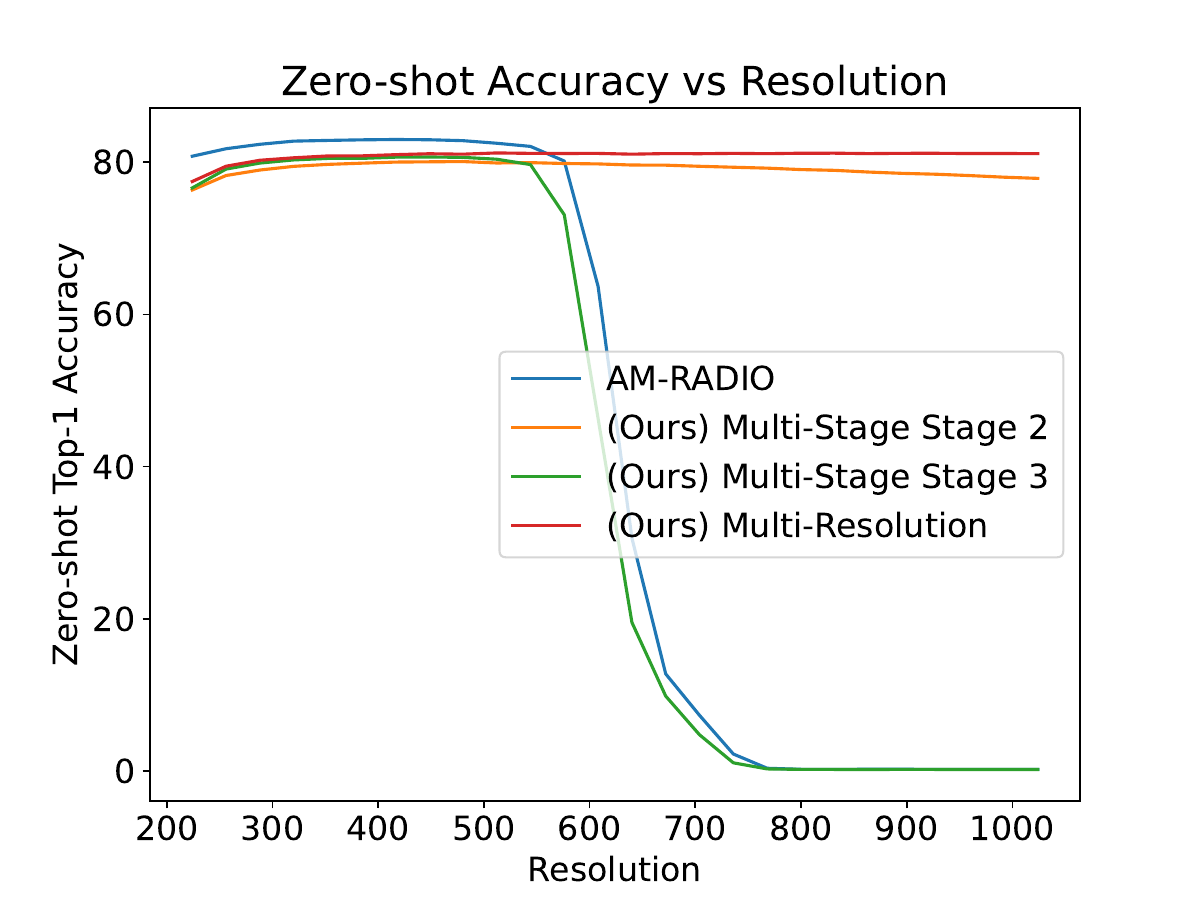}
  \includegraphics[width=\linewidth]{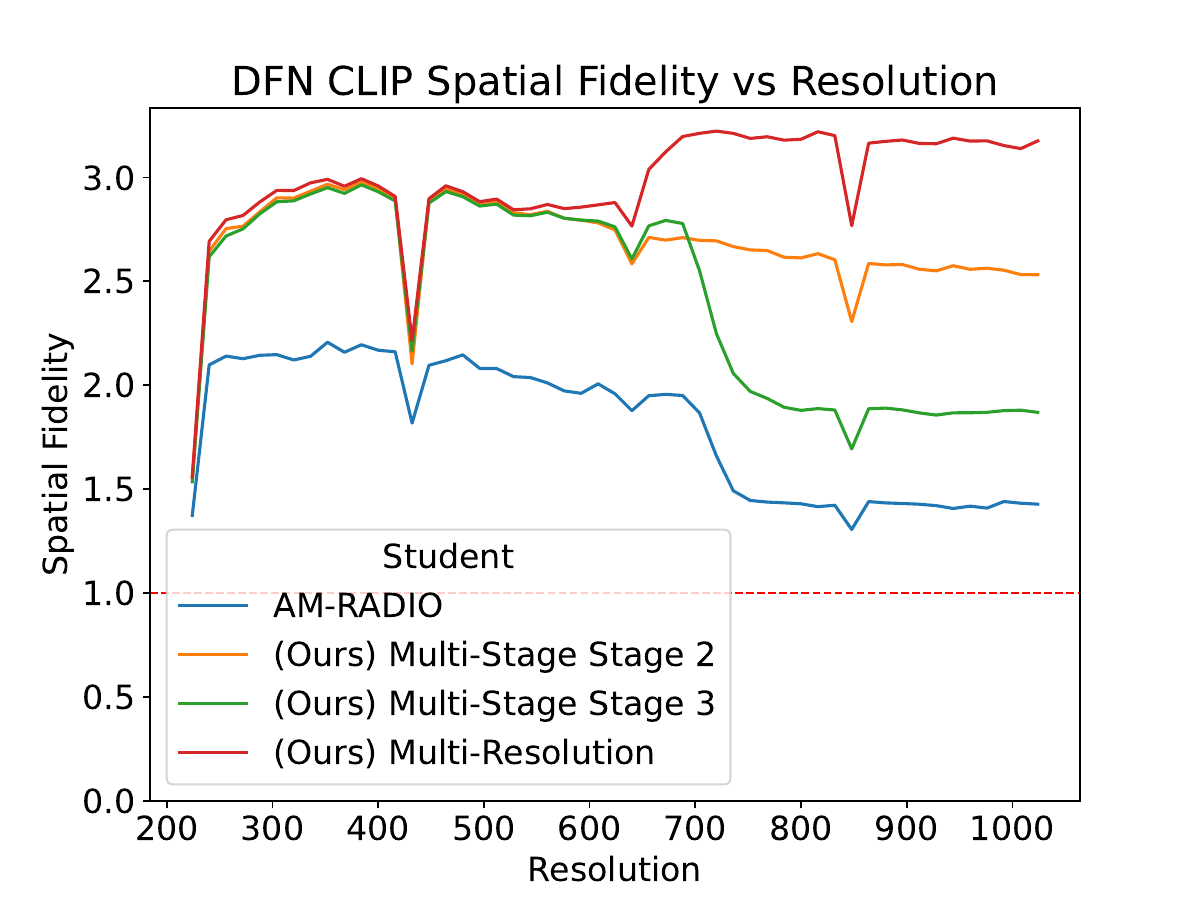}
  \includegraphics[width=\linewidth]{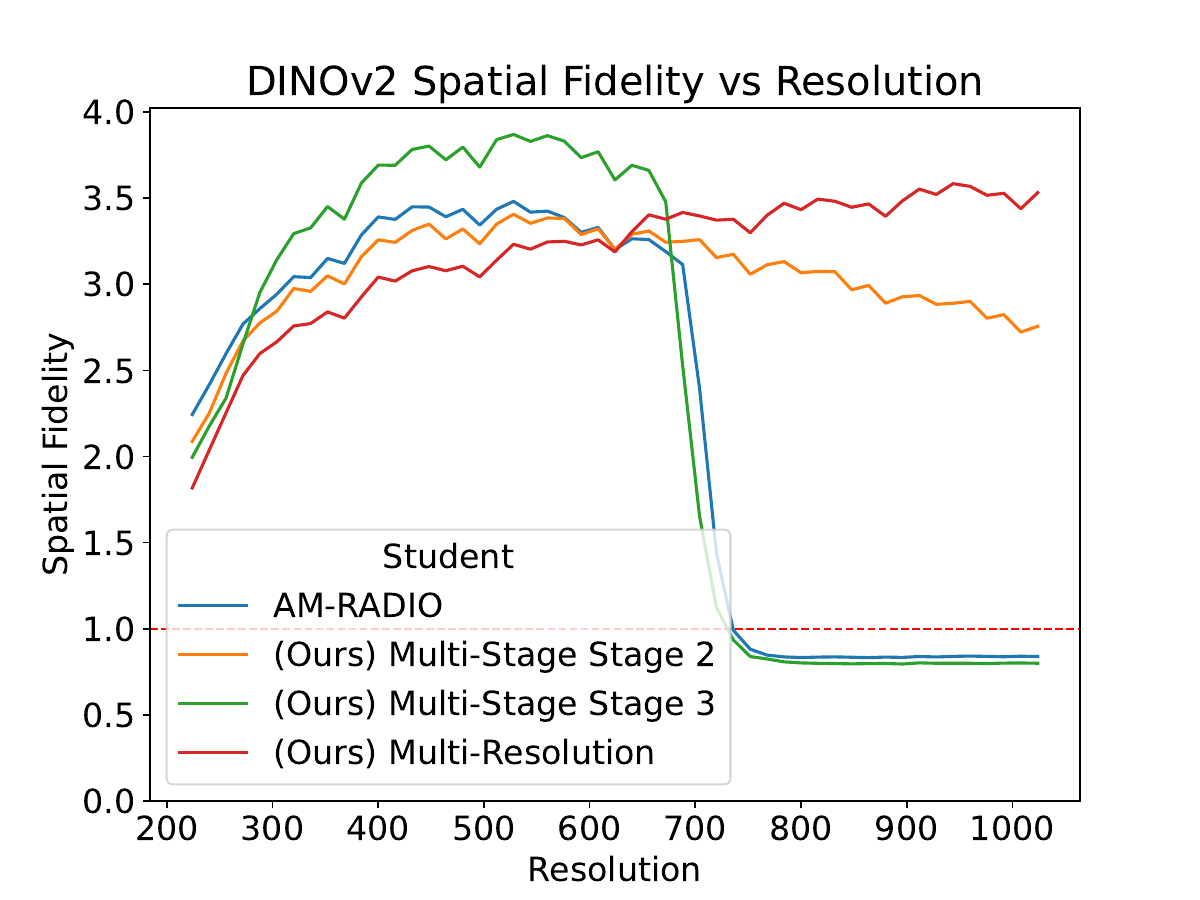}
  \includegraphics[width=\linewidth]{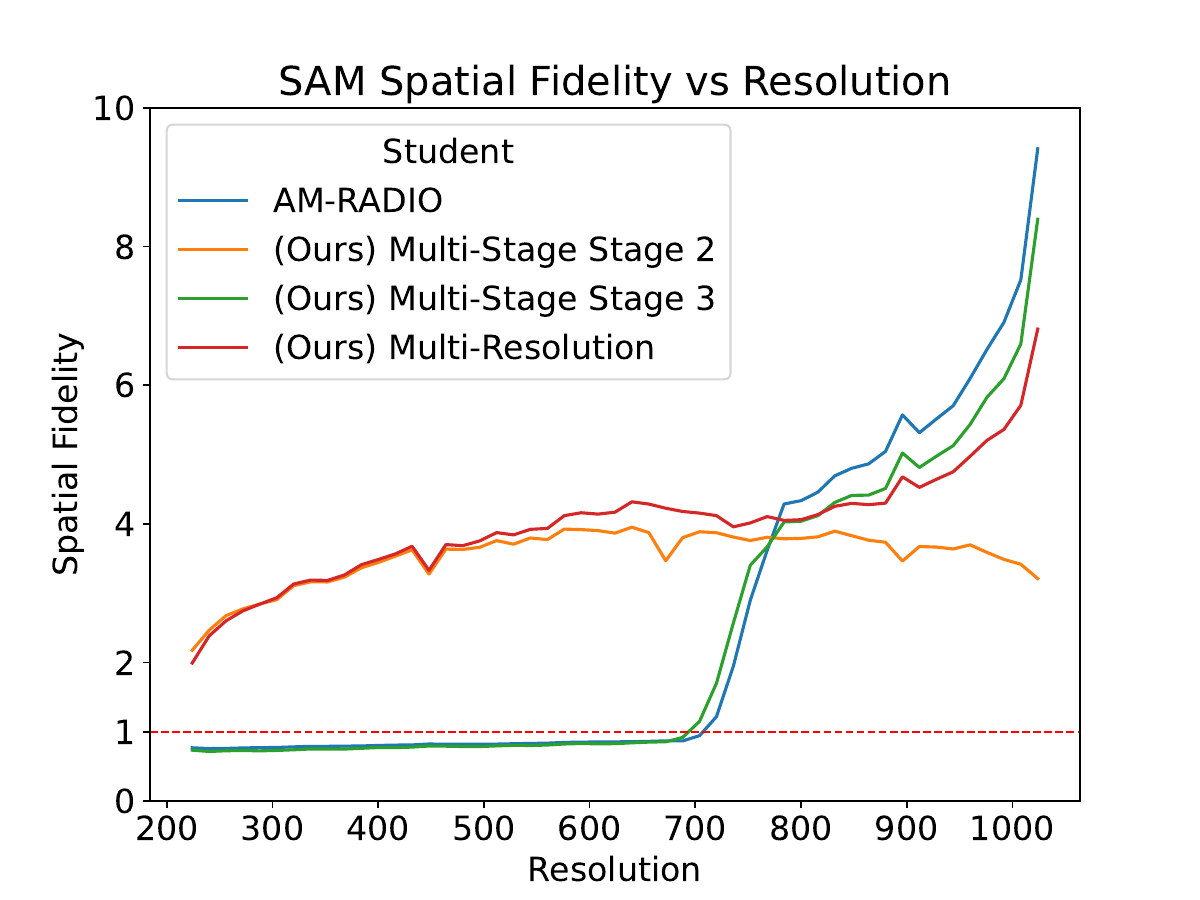}
  }
  \caption{Zero-Shot Accuracy and teacher matching fidelity metrics as a function of input resolution. AM-RADIO (aka RADIOv2.1, config \B{$\mathcal{A}$}), and our ``Multi-Stage Stage 3'' model exhibit mode switching behavior. ``Multi-Stage Stage 2'' doesn't exhibit the behavior, owing to all teachers being trained solely at low resolution, however quality at high resolution quickly degrades. ``Multi-Resolution'' (config \B{$\mathcal{B}$}) training fixes the mode switch, and allows the model to be strong across a larger range of resolutions compared to Stage 2.}
  \label{fig:zero-shot-accuracy-vs-resolution}
\end{figure*}

\begin{figure}
  \centering
  \includegraphics[width=\linewidth]{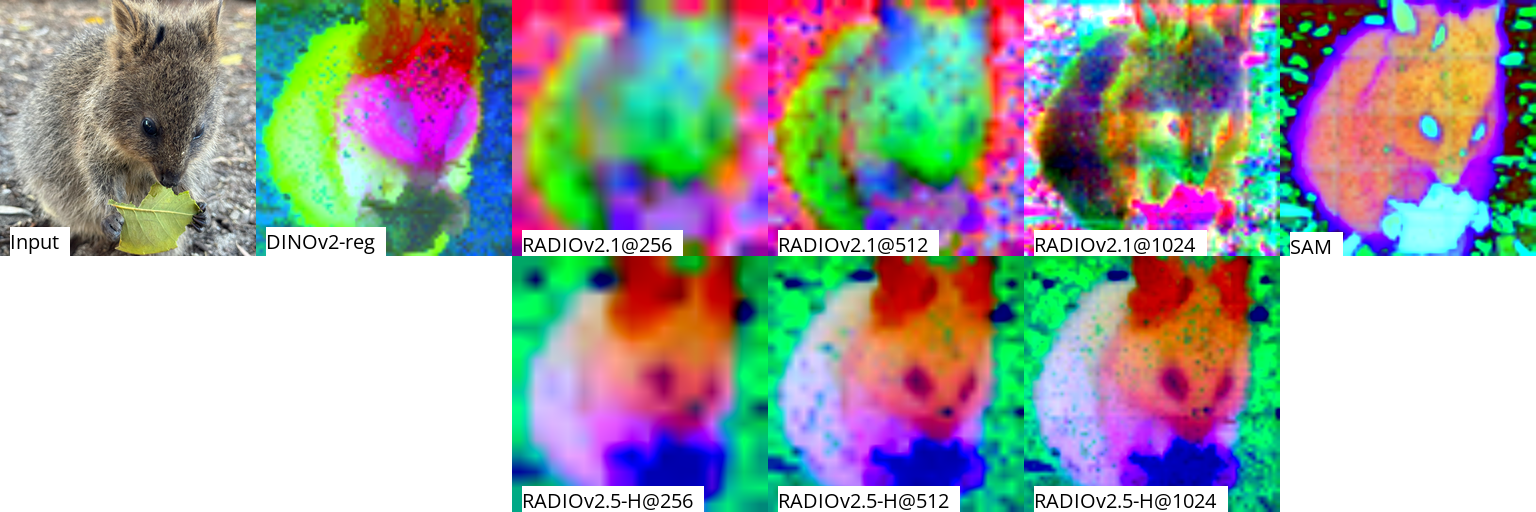} \\   
  \caption{Visualizations of patch token features across input resolutions. We use PCA to map patch tokens to RGB colors. \B{Top}: Input image, followed by visualizations of DINOv2, RADIOv2.1 (config \B{$\mathcal{A}$}) at $256^2$, $512^2$, $1024^2$ resolutions, and SAM. \B{Bottom}: RADIOv2.5-H (config \B{$\mathcal{D}$}). As resolution increases, RADIOv2.1 features resemble SAM’s, even showing VitDet windowing artifacts, while RADIOv2.5 remains more self-consistent and less noisy.}
  \label{fig:mode-switch-features}
\end{figure}

Although AM-RADIO \cite{ranzinger2024radio} supports a wide range of input resolutions, we noticed poor performance on high-resolution benchmarks. As illustrated in Figure~\ref{fig:zero-shot-accuracy-vs-resolution}, Zero-Shot Top-1 accuracy drops sharply after inputs are larger than roughly 512px. This showcases an undesirable property of this model, which is that it doesn't reliably scale to high-resolution images. AM-RADIO hints at this issue in its conclusion: the model exhibits a ``mode switching'' behavior as resolution increases. This prompted us to visualize the features and investigate the effect of a resolution increase. Figure~\ref{fig:mode-switch-features} illustrates the issue well: at resolutions lower than or equal to $512^2$, the features most closely resemble those of DINOv2 and appear to be expressive of depth and semantic content. At higher resolutions, the model starts to behave more like SAM, with features that show sharp delineations around contours, with homogeneous contents within. This behavior is intuitive given the fact that in the high-resolution regime the student only sees SAM features. Table~\ref{tab:scale-variance} shows a quantified measure of feature variance across scales (see implementation in appendix). DINOv2 exhibits much smaller scale variance.


\subsection{Token Count}

Dense downstream tasks, such as pixel-level semantic segmentation and depth estimation, benefit from receiving fine features from a vision encoder. However, for vision-language models like LLaVA~\cite{liu2024llavanext}, where visual tokens are integrated into a sequence of text embeddings, an excessive number of vision tokens can negatively impact performance or lead to sequence overflows. 
Pixel unshuffling, as proposed in InternVL1.5~\cite{chen2024far}, addresses this by reducing the number of vision tokens, grouping $2 \times 2$ tokens along the channel dimension. However, this approach cannot adapt to varying information densities within an image. For instance, a text document might have large areas of white background that should be more strongly compressed than areas containing text.
 
\section{Method}\label{sec:method}

\subsection{Scale Equivariance}

\subsubsection{A Measure of Scale Equivariance}

We define a measure of scale equivariance for a set of multi-scale features by normalizing and interpolating all features to the scale of the lowest resolution, then calculating the spatial average of the variance along the scale dimension. Our formula can be found in appendix~\ref{apdx:scale_eq}.

As can be seen in Table~\ref{tab:scale-variance}, our baseline student exhibits much worse scale equivariance than that of DINOv2 (the only teacher that is capable of operating at multiple resolutions, hence our only point of comparison).

\subsubsection{Tiling}

Tiling is commonly employed in VLMs (LLaVa-NeXT\cite{liu2024llavanext}, InternVL1.5\cite{chen2024far}) as a way to enable high-resolution inference when the vision encoder only supports fixed-resolution inputs. Thus it could be argued that scale equivariance is of little importance, given that it is possible to resort to tiling instead of increasing input image resolution. However tiling incurs additional challenges:

\begin{itemize}
    \item Tiling makes resolution increases very coarse.
    \item As the number of tiles increases, the vision encoder sees an increasingly small subset of the full image, limiting its ability to reason about the full context, or even know what it's looking at entirely.
\end{itemize}

\begin{figure}[ht]
    \centering
    \includegraphics[width=\linewidth]{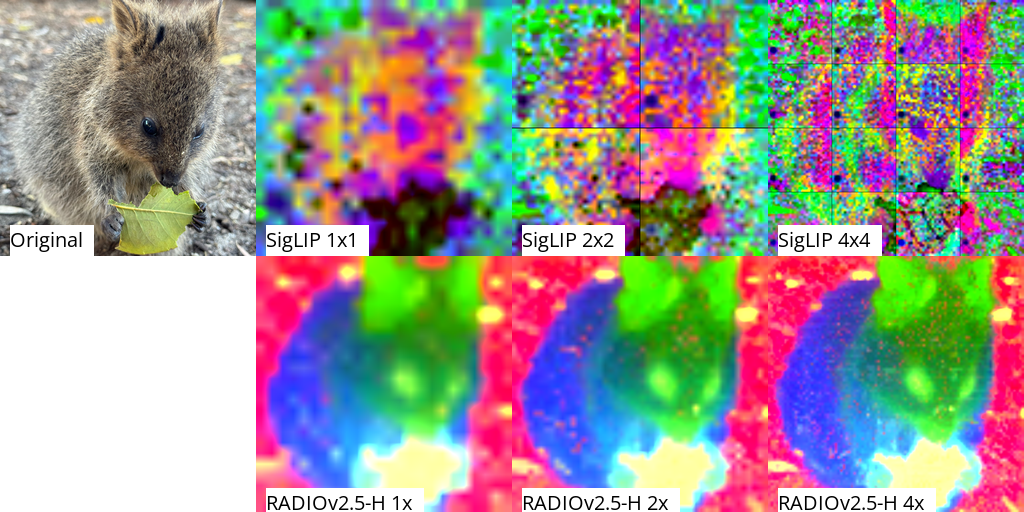} 
    \caption{Qualitative comparison of tiling and high-resolution inference. Tiling emulates high-res SigLIP inference. \textbf{Top:} From left to right: input, SigLIP features using 1, $2 \times 2$ and $4 \times 4$ tiles. \textbf{Bottom:} Single-pass inference using RADIOv2.5 (ours) at equivalent image sizes. SigLIP features vary significantly across different tiling arrangements.}
    \label{fig:tiling}
\end{figure}

In Figure~\ref{fig:tiling} we use SigLIP~\cite{zhai2023sigmoidlosslanguageimage}, a popular choice of Vision Encoder (Cambrian~\cite{tong2024cambrian1fullyopenvisioncentric}), and show its features for multiple tiling arrangements. We notice qualitative inconsistency in the way features get represented at multiple scales. We can also apply our measure of scale equivariance, albeit rather coarsely, to quantify this inconsistency (Table~\ref{tab:scale-variance}). In both cases we notice the detrimental effect of tiling on scale equivariance. Tiling also incurs a substantial increase in the number of vision tokens, by a factor of up to $12 \times$~(\cite{chen2024far}), with immediate effects on VLM latency.

\findingbox{High-resolution inference through tiling causes the vision encoder to lose global context and exhibit poor scaling equivariance.}

\subsection{Multi-Resolution Training}
\label{sec:mulires}

The previous observations motivate a change in the training schedule: with multi-resolution training we enable the student to learn from all teachers across multiple resolutions. This is easily achieved with DINOv2, since DINOv2 can infer images at any resolution. For CLIP teachers, we feed them with images at the teacher's native resolution, and we feed the student with images at multiple resolutions. We interpolate student features down to the resolution of the teacher's features before applying the loss function. SAM presents an extra challenge, as noted in AM-RADIO\cite{ranzinger2024radio}, since interpolating SAM features significantly deteriorates their quality. Therefore in order to train our student against SAM at lower resolutions than $1024^2$, we pad smaller images to $1024^2$, then crop SAM features to the effective size of the unpadded image.

Training the student for 600k iterations with a multi-resolution setup is costly. Thus we break down training into three stages:

\begin{itemize}
    \item In a \textit{first stage}, we train the student at low resolution ($256^2$) for 300k iterations.
    \item In a \textit{second stage}, we train the student at medium resolution ($432^2$) for 300k iterations.
    \item In the \textit{third stage}, we train the student simultaneously at $432^2$ and $1024^2$ resolutions for 300k iterations.
\end{itemize}

\begin{figure}
  \centering
  \includegraphics[width=\linewidth]{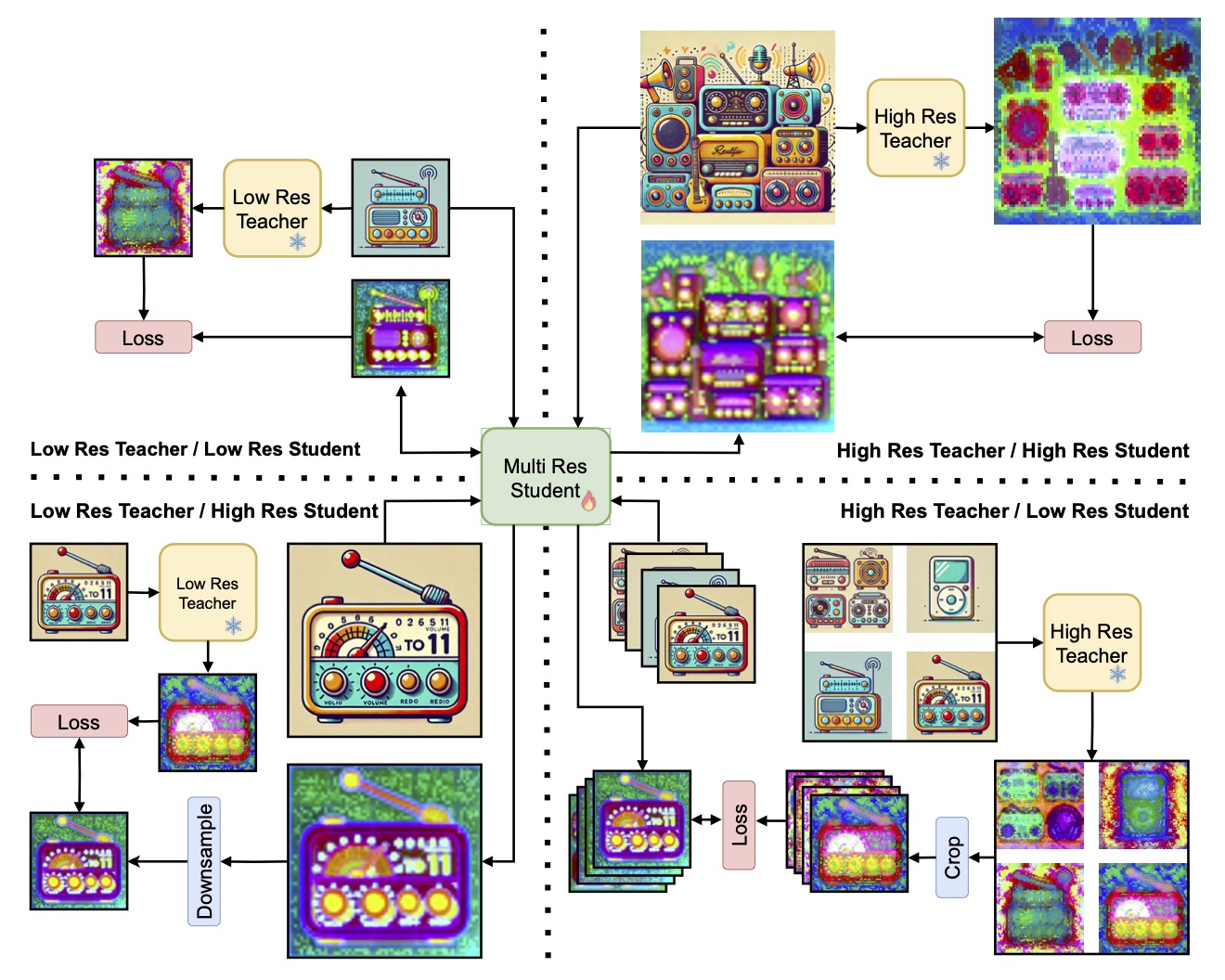} \\   
  \caption{Illustration of the multi-resolution training setup. The student learns from all teachers at all resolutions. High-res student features are downsampled to match the resolution of low-res teachers. High-res teachers are provided with mosaics of smaller images and their output features are cropped out in order to train a low-res student.}
  \label{fig:multires}
\end{figure}

Multi-resolution training is illustrated in Figure~\ref{fig:multires}, which depicts four training regimes based on whether the teachers and/or students process low- or high-resolution images. When both the student and teachers use either low- or high-resolution images, no particular challenge arises. However, training a high-resolution student against a low-resolution teacher requires downscaling the student's features to match those of the teacher. Conversely, training a low-resolution student against a high-resolution teacher necessitates a careful augmentation technique, as described in Section~\ref{sec:mosaic-augmentation}.

\findingbox{For the student model to be consistently accurate across resolutions,  it is sufficient to match all teachers at all resolutions, and to train at two resolutions simultaneously in the final training stage.}

\begin{table}[]
    \centering
    \centering\resizebox{0.8\linewidth}{!}{
     
    \begin{tabular}{lccc}
    
    \toprule
     & \multicolumn{2}{c}{Equivariance $\downarrow$} \\
        \textbf{Model}       & \textbf{Fine Scale } & \textbf{Coarse Scale} \\
        \midrule        
        DINOv2-g-reg    & 0.126    & 0.178 \\        
        AM-RADIO-L         & 0.357    & 0.476 \\        
        RADIOv2.5-B     & \B{0.102}    & 0.168 \\
        RADIOv2.5-L     & \B{0.102}    & \B{0.165} \\
        RADIOv2.5-H     & 0.119    & 0.193 \\        
        \midrule
        RADIOv2.5-L (tiling) & N/A & \B{0.369} \\
        SigLIP (tiling)      & N/A & 0.623 \\        
    \bottomrule
    \end{tabular}
    }
    
    \caption{Scale equivariance, computed over a small dataset of natural images (shown in Appendix). \textbf{Fine:} we scale image resolutions from $384^2$ to $1568^2$ by increments of the model's patch size. \textbf{Coarse:} we scale images from $384^2$ to $1920^2$ by multiples of $384^2$. For the models that are denoted as ``tiling'', we perform inference over tiles of $384^2$) and re-assemble features to produce the high-resolution features.
    }
    \label{tab:scale-variance}
\end{table}

\subsection{Mosaic Augmentation}
\label{sec:mosaic-augmentation}

The training schedule described in Section \ref{sec:mulires} involves running SAM inference on padded images, using cropped features to train the student against SAM at low resolution. This approach incurs a substantial computational cost, as SAM is the most resource-intensive teacher.

\begin{figure}[h]
    \centering
    \includegraphics[width=0.45\columnwidth]{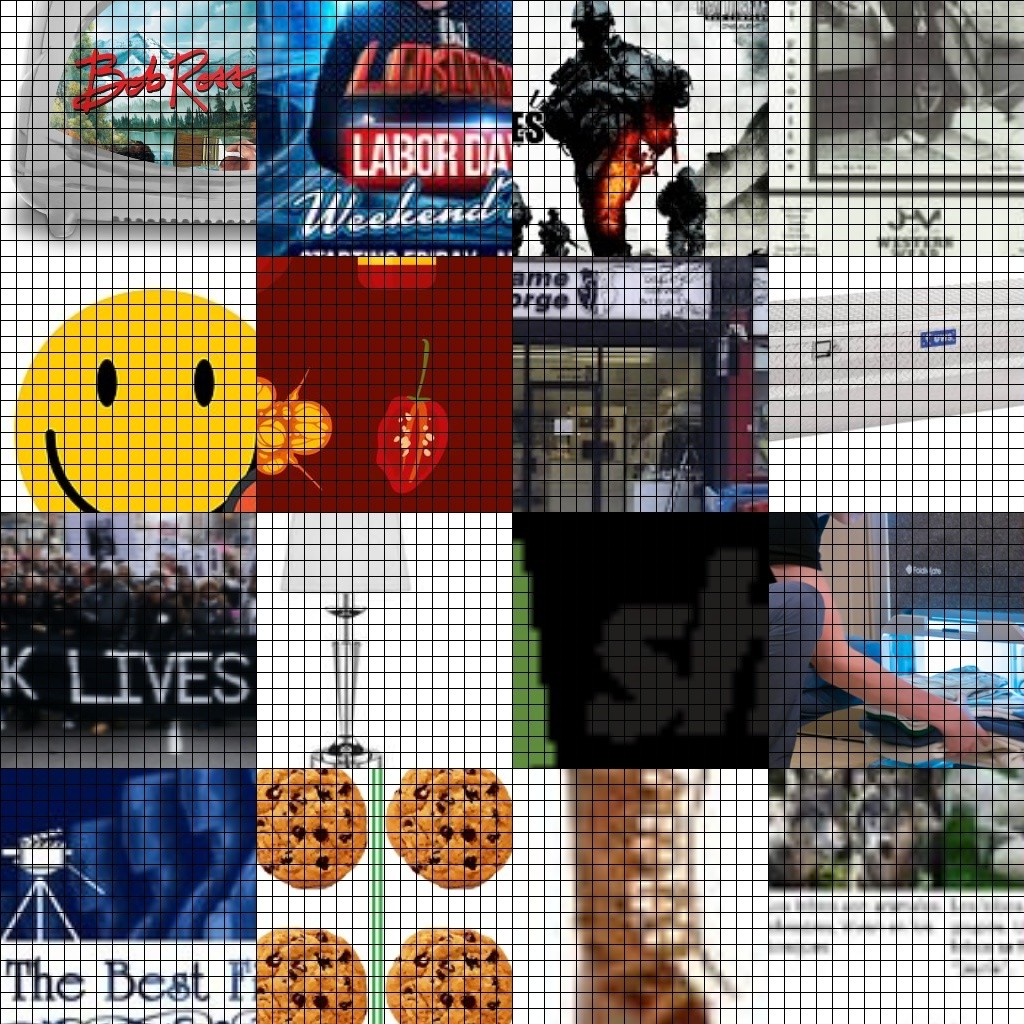}
    \hfill
    \includegraphics[width=0.45\columnwidth]{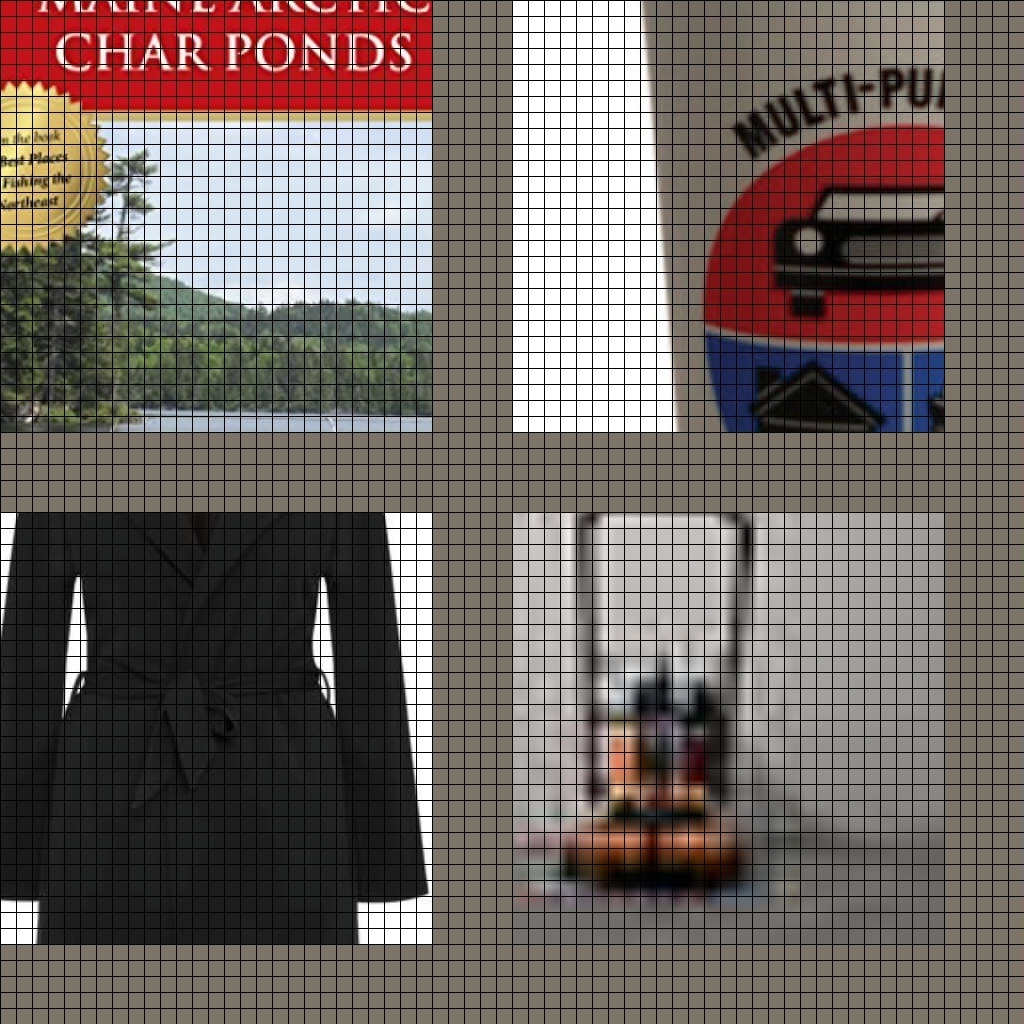}
    \caption{Sample mosaic augmentations. Grid lines represent $16^2$ patches (only shown here for visualization purposes). \textbf{Left:} A $4 \times 4$ arrangement with a $1024^2$ image comprising $16 \times 256^2$ sub-images. \textbf{Right:} A $2 \times 2$ arrangement comprising $4 \times 432^2$ sub-images, with 80 pixels of padding around each.}
    \label{fig:mosaic-augmentations}
\end{figure}

To improve efficiency when training a student at a resolution $\leq 512^2$ against SAM, we can instead create a mosaic of $k \times k$ images, with $k = \left\lfloor \frac{1024}{x} \right\rfloor$ and $x$ being the student resolution, resulting in a single $1024^2$ image. We then perform SAM inference on this mosaic and extract $k^2$ individual feature maps to train the student. Mosaic augmentation includes padding as needed to maximize efficiency. For example, to train a student at $432^2$ resolution, we can create a $2 \times 2$ mosaic with 80-pixel padding around each image. Figure \ref{fig:mosaic-augmentations} shows sample mosaic augmentations under 256 and 432 student resolutions. Qualitatively, we observe cleaner features after applying mosaic augmentation, which we believe is due to the increased diversity in image positions, helping to reduce positional encoding artifacts. We also show SAM's PCA features for these mosaic images in figures \ref{fig:mosaic-and-features-4x4} and \ref{fig:mosaic-and-features-2x2}.

\findingbox{Mosaic augmentation greatly reduces the training cost associated with learning from high-resolution teachers and eliminates the need for feature interpolation. Student quality is even \textit{improved} with this optimization.}

\subsection{Teacher Loss Balancing}\label{sec:phi-s}

PHI-S \cite{ranzinger2024phisdistributionbalancinglabelfree} highlights the significant variations in activation magnitudes among vision foundation models, observing that SAM's activations tend to overshadow those of CLIP and DINOv2 models. We adopt the PCA-Hadamard Isotropic Standardization (PHI-S) method, achieving improved balance among teacher losses. PHI-S rotates teacher activations to evenly distribute variance across all channels and then scales them to obtain unit variance. This process can be easily reversed by projecting the student activations back into each teacher’s original feature space, or even by modifying the final linear projection of the adaptor MLP after training. PHI-S enhances training stability and overall benchmark performance. For a given teacher feature map $\mathbf{X}$ with embedding size $C$, PHI-S applies the following transformation:

\begin{equation}
    \mathbf{X}_i' = \phi_i^{-1} \mathbf{R}_i\mathbf{X}_i,\quad
   \mathbf{R}_i = \mathbf{H}_C \mathbf{U}_i^\intercal,\quad
   \phi_i = \sqrt{\frac{1}{C} \sum_j^C \lambda_j}
\end{equation}

Where $H_C$ is a normalized Hadamard matrix of dimension $C$, $\lambda_j$ are the Eigen values of the covariance matrix $\cov{\mathbf{X}}$, and $\mathbf{U}$ are the corresponding eigenvectors. $\phi_i$ and $\mathbf{R}_i$ are specific to the $i$th teacher.

As a starting point, we define a measure of fidelity, similar to that in the classification-distillation literature (\cite{stanton2021does,beyer2022patient}), however we do so without the use of labels or explicitly produced distributions over classes. Instead, since our loss objective is to directly match the features of the teachers, we have

\begin{equation}
\begin{split}
    F_i[X] &= \frac{\var\left[ t_i(X) \right]}{\var\left[f(X) - t_i(X)\right]} = \frac{\phi_i^2}{\mse(f(X), t_i(X))}
\end{split}
    \label{eq:fidelity}
\end{equation}

with $f(X)$ being the student feature distribution, and $t_i(X)$ being the $i$th teacher distribution. This function represents the ratio of the target distribution variance to the student's estimation error variance. A value of $\leq 1$ means random sampling from the teacher distribution would be better, and $\infty$ would be perfect matching. We show the results of this in table \ref{tab:fidelity}, where it is apparent that the baseline allocates too much energy to matching SAM due it its disproportionately large distribution variance, consistent with \cite{ranzinger2024phisdistributionbalancinglabelfree}.
The errors relative to the variance are overall smaller when applying PHI-S.

\begin{table}[]
    \centering\resizebox{\linewidth}{!}{
    \begin{tabular}{lccccc}
        \hline
        \multirow{2}{*}{Teacher} & \multirow{2}{*}{$\phi^2$} & \multicolumn{2}{|c|}{MSE} & \multicolumn{2}{|c|}{F[X]} \\
        & & Baseline & PHI-S & Baseline & PHI-S  \\
        \hline
        DFN CLIP    & 5.831E-4 & 5.100E-4 & 2.418E-4 & 1.143      & \bf{2.411} \\
        OpenAI CLIP & 0.820    & 0.570    & 0.525    & 1.438      & \bf{1.563} \\
        DINOv2-g    & 1.729    & 0.222    & 0.206    & 7.799      & \bf{8.377} \\
        SAM         & 27.263   & 3.719    & 5.313    & \bf{7.331} & 5.132 \\
        \hline
    \end{tabular}
    
    }
    \resizebox{0.6\linewidth}{!}{
    \begin{tabular}{lcc}
        \multirow{2}{*}{Geometric Mean} & Baseline & PHI-S \\
         & 3.114 & \bf{3.568} \\
         \hline
    \end{tabular}
    }
    
    \caption{Fidelity results for each teacher, comparing between the baseline and PHI-S. Higher values are better.}
    \label{tab:fidelity}
\end{table}

\findingbox{PHI Standardization helps balance the energy spent learning from each teacher.}

\subsection{Is SAM a good teacher?}
SAM~\cite{kirillov2023sam} has been a controversial choice in the recent agglomerative models literature. AM-RADIO~\cite{ranzinger2024radio} struggled to prove that its inclusion improved any metrics. Theia~\cite{shang2024theiadistillingdiversevision} specifically ablated whether to include SAM, and found that it degraded their metrics. UNIC~\cite{sariyildiz2024unicuniversalclassificationmodels} opted as well not to include SAM. Based on the findings with PHI-S~\cite{ranzinger2024phisdistributionbalancinglabelfree}, and our confirmation of imbalance in section \ref{sec:phi-s}, it seems that a major problem with SAM may just have been that interpolating its features is a really bad thing, and also that the distribution was extremely unbalanced; something that PHI-S corrects. We chose to re-run the study of whether SAM is a good teacher now that we have a new bag of tricks. In particular, we run the first two stages of multi-stage training, we use the mosaic augmentation for SAM in both of these stages, and we apply PHI-S to all teachers. We show the results in table \ref{tab:sam_inclusion} where it is clear that including SAM has negligible (but positive) impact on our classification benchmarks, and strong positive effects on dense tasks such as semantic segmentation and 3D probing. SAM's inclusion also enables novel opportunities such as those found in VideoSAM~\cite{guo2024videosamopenworldvideosegmentation} where they employ AM-RADIO's SAM adaptor and backbone simultaneously for video segmentation.

\begin{table}[]
    \centering
    \resizebox{\linewidth}{!}{
    \begin{tabular}{l|cc|c|cccc}
         \hline
         Variant & Zero Shot & kNN & ADE20k & Depth & SNorm & MultiView & SPairs \\
         \hline
         \multicolumn{8}{c}{Stage 1} \\
         \hline
         No SAM & \bf{79.38} & 83.17 & 50.27  & 82.54 & 61.03 & 58.12 & 51.97 \\
         SAM    & 79.37 & \bf{83.29} & \bf{51.14} &  \bf{82.60} & \bf{61.88} & \bf{58.91} & \bf{52.49} \\
         \hline
         \multicolumn{8}{c}{Stage 2} \\
         \hline
         No SAM & 80.43 & 83.83 & 50.24 & \bf{83.29} & 61.43 & 58.98 & \bf{54.72} \\
         SAM    & \bf{80.47} & \bf{83.92} & \bf{51.36} & 83.17 & \bf{62.80} & \bf{61.36} & 54.66 \\
         \end{tabular}}
    \caption{Ablation on whether to include SAM in the teacher set for the first two stages of multi-stage training.}
    \label{tab:sam_inclusion}
\end{table}

\findingbox{All teachers are beneficial, including SAM, despite recent trends. It also has broad downstream applicability, granting our student the same abilities.}

\subsection{Partitioning}\label{sec:partitioning}
In PHI-S~\cite{ranzinger2024phisdistributionbalancinglabelfree}, the authors opted to put teachers in their own partitions, which reduces the teacher overhead (as the per-teacher batch size is reduced). However, the paper does not ablate whether this choice came with model quality consequences. In Table \ref{tab:partitioning} we study the number of partitions for the first two stages of training. We find that fewer partitions is strongly better for summarization tasks, and less clear for dense tasks.

\begin{table}[]
    \centering
    \resizebox{\linewidth}{!}{
    \begin{tabular}{c|cc|c|cccc|c}
         \hline
         \# & Zero Shot & kNN & ADE20k & Depth & SNorm & MultiView & SPairs & SAM \\
         \hline
         \multicolumn{8}{c}{Stage 1} \\
         \hline
         1 & \bf{79.37} & \bf{83.29} & \bf{51.14}  & 82.60 & \bf{61.88} & \bf{58.91} & 52.49 & \bf{71.51} \\
         2 & 78.70 & 83.00 & 50.88  & 82.32 & 60.78 & 58.49 & 52.32 & 71.29 \\
         4 & 77.93 & 82.61 & 51.09 & \bf{82.89} & 61.26 & 58.47 & \bf{52.54} & 71.02 \\
         \hline
         \multicolumn{8}{c}{Stage 2} \\
         \hline
         1 & \bf{80.47} & \bf{83.92} & \bf{51.36}  & \bf{83.17} & \bf{62.80} & \bf{61.36} & 54.66 & \bf{73.62} \\
         4 & 78.82 & 83.41 & 51.34  & 82.99 & 61.58 & 59.80 & \bf{55.96} & 73.03 \\
    \end{tabular}
    }
    \caption{Ablation on number of partitions for the first two stages of multi-stage training.}
    \label{tab:partitioning}
\end{table}

\findingbox{Minimizing the number of partitions seems to be beneficial, assuming you can afford the teacher overhead. Under compute constraints, partitioning is an effective strategy to reduce the overhead.}

\subsection{SigLIP Teacher}

Building on previous work (\cite{tong2024cambrian1fullyopenvisioncentric}, \cite{lin2023vila}), we replace OpenAI-CLIP~\cite{radford2021learningtransferablevisualmodels} with SigLIP~\cite{zhai2023sigmoidlosslanguageimage}, defining this as our configuration \B{$\mathcal{C}$}. Our choice is validated by the significant improvements observed in VLM tasks, as shown in Table~\ref{tab:ablation_table}.

\subsection{Token Compression}\label{sec:token_compression}

\begin{figure}
    \centering
    \includegraphics[width=\linewidth]{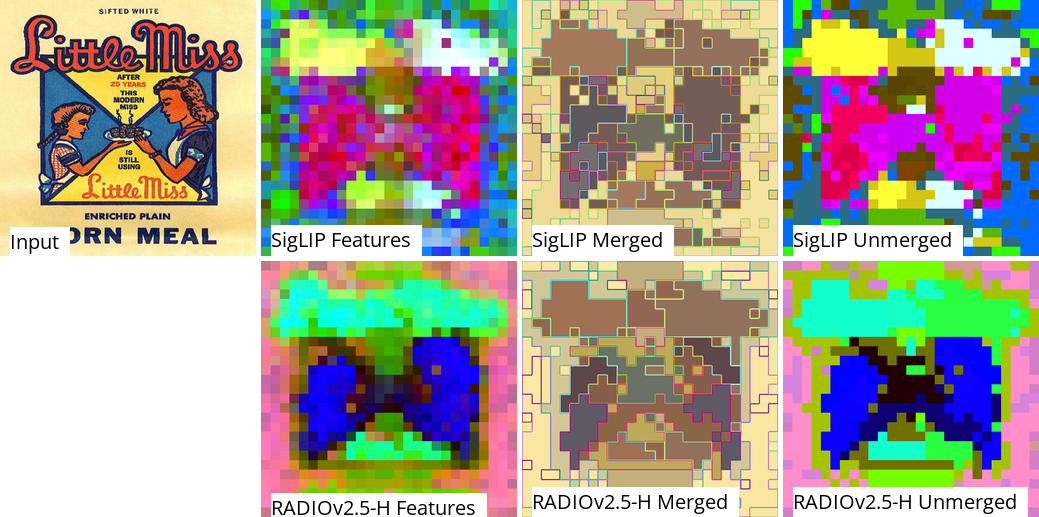}
    \caption{Visualizations of SigLIP and RADIOv2.5-H model features before and after ToMe compression. From left to right: input image; PCA visualization of features; merged tokens; PCA visualization of unmerged tokens. RADIOv2.5 is better able to compress background regions, leaving more tokens available for semantically rich features.}
    \label{fig:tome}
\end{figure}

\begin{figure}[t]
  \centering
  \includegraphics[width=\linewidth]{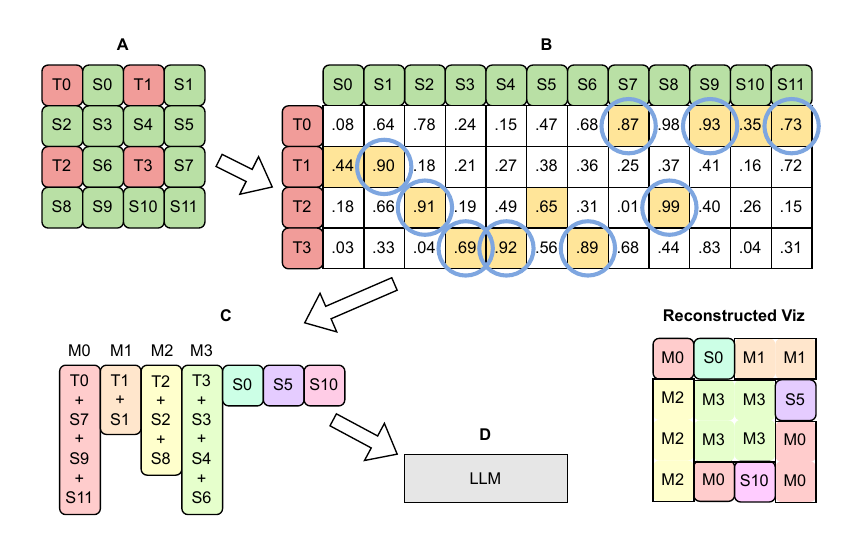}
  \caption{Illustration of bipartite matching~\cite{bolya2023tokenmergingfaststable} using $2 \times 2$ strided pattern with $r=9$. \textbf{A}: The original tokens, with the red `T\#' squares being assigned as targets, and the green `S\#' squares being assigned as sources. \textbf{B}: We compute the affinity between each source and target. We only consider the maximum affinity for each source (shown in yellow). We find the $r$ highest affinity yellow squares, and merge those into their respective targets. \textbf{C}: The output tokens with merged values when a given `T\#' was assigned one or more sources. \textbf{D}: The final $7$ tokens are fed to the LLM. \textbf{Reconstructed Viz}: From \textbf{(C)}, we can visualize the compressed original feature map by broadcasting the merged tokens to all of the source locations.
  }
  \label{fig:bipartite-matching}
\end{figure}

Inspired by ToMeSD\cite{bolya2023tokenmergingfaststable}, we evaluate the use of bipartite soft matching to merge similar tokens. We apply strided partitioning to ensure that each image region retains some representation in the compressed features. For evaluation, we track merged token indices, enabling us to unmerge tokens and measure the reconstruction error and inform hyperparameter selection. The method is illustrated in Figure~\ref{fig:bipartite-matching}.

While ToMe recommends using attention keys for matching, we found that using the features directly reduces reconstruction error. Table~\ref{tab:tome-reconstruction-error} in the appendix summarizes our findings. Another slight departure from ToMe’s recommendation is that we apply merging at the output of the vision encoder rather than incrementally. As shown in Table~\ref{tab:token-merging-inline-vs-output}, this approach yields an average improvement of +1.4 on VLM benchmarks.

We also qualitatively evaluate token merging by visualizing feature and token merging outcomes in Figure \ref{fig:tome}. Token merging is assessed within the context of VLMs, with results reported in Table \ref{tab:ablation_table}.

We further note that token merging can be applied to other vision encoders and demonstrate its application on SigLIP. Interestingly, token merging yields a greater relative improvement on RADIOv2.5 than on SigLIP, a finding that aligns with the feature visualizations in Figure \ref{fig:tome} and the measured reconstruction error in Table \ref{tab:tome-reconstruction-error}. We hypothesize that SigLIP’s higher feature noise makes clustering more error prone.

\begin{table}[]
    \centering
    \centering\resizebox{\linewidth}{!}{
    \begin{tabular}{lcccccccccc}
    
    \toprule
        \textbf{Res}  & \textbf{Method} & \rotatebox{90}{\textbf{Tokens/im}}  & \rotatebox{90}{\textbf{TextVQA}} & \rotatebox{90}{\textbf{ChartQA}} & \rotatebox{90}{\textbf{DocVQA}} &  \rotatebox{90}{\textbf{InfoVQA}} & \rotatebox{90}{\textbf{OCRBench}} \\
        \midrule        
\multirow{2}{*}{512} &  $2^2$ Unshuffle & 256 & 65.0 & 23.7 & 45.7 & 32.9 & 40.2 \\
    & ToMe r=768 & 256 &  \textbf{65.7} & \textbf{25.8} & \textbf{49.7} &  \textbf{34.9} & \textbf{41.2} \\
        \midrule
\multirow{2}{*}{768} & $3^3$ Unshuffle & 256 &  65.8 & 25.9 & 49.7 & 35.2 & 40.9\\
    & ToMe r=2048 & 256 & \textbf{69.7} & \textbf{30.4} & \textbf{52.3}   & \textbf{36.2} & \textbf{42.9} \\
        \midrule
\multirow{2}{*}{1024} & $4^4$ Unshuffle & 256 & 66.1 & 23.8 & 46.6 &  34.1 & 38.9 \\
    & ToMe r=3840 & 256 & \B{69.3} & 28.1 & 49.1 & 34.6 & 42.7  \\
    & ToMe r=3584 & 512 & 68.9 & \B{31.3} & \B{53.9} & \B{37.0} & \B{44.6}  \\
    \bottomrule
     \end{tabular}
    }        
    \caption{VLM benchmarks across various compression methods. All configurations produce 256 vision tokens per image, except for the last row (512 tokens). Token Merging outperforms Pixel Unshuffling on all benchmarks.}
    \label{tab:token-compression-vila}
\end{table}

\findingbox{Token Merging is very effective at retaining the most diverse information under high compression ratios.}

\subsection{Feature Selection}

For each image, our foundation model outputs a summary vector along with patch tokens at a granularity of one per $16^2$ input pixel block. For image-level tasks such as classification, search, or curation, the summary vector provides a rich embedding. For dense tasks, such as segmentation or 3D understanding, the patch tokens are a natural choice. However, as demonstrated in previous work, incorporating additional intermediate activations further enhances performance. For example, \cite{elbanani2024probing} uses a Dense Prediction Transformer (DPT) head \cite{ranftl2021visiontransformersdenseprediction} for 3D reasoning, while \cite{yao2024denseconnectormllms} averages multiple ranges of intermediate activations to feed into an LLM for VLMs.

In this section, we investigate various feature selection methods and present the results in Table \ref{tab:dense-features-semseg}. We experiment with \textit{sparse} feature selection (selecting activations from individual layers throughout the model) and \textit{dense} feature selection (aggregating information across all layers by averaging groups of layer activations). We examine the impact of feature selection in conjunction with different downstream heads (linear or DPT probe).

Our findings show that a linear probe alone is insufficient to leverage additional information from intermediate layer activations. However, when a DPT head is employed, it effectively incorporates this additional information. We note, however, that it is challenging to disentangle the benefits provided by additional feature information, and those of the extra learnable parameters of the DPT head. Unlike \cite{yao2024denseconnectormllms}, we do not observe a positive impact of using dense features in VLMs.

\findingbox{Intermediate layer activations greatly benefit downstream tasks if a non-linear transformation is employed.}

\begin{table}[]
    \centering
    \centering\resizebox{\linewidth}{!}{
     
    \begin{tabular}{lcccccc}
    
    \toprule
        \B{Layers}  & \B{Aggre-} & \B{Head} & \B{ADE20k} & \B{Depth} & \B{Surf} & \textbf{Overall} \\
          & \B{gation} && &  & \B{Normals} &  \\
        \midrule        
        31 & N/A & Linear & 52.47 & 82.9 & 57.0 & 61.215 \\         
        7-15-23-31 & Sparse & Linear & 52.99 & 82.5 & 59.6 & 62.03 \\
        (0-9)-(10-19)-(20-30)-31 & Dense & Linear & 52.96 & 82.7 & 59.5 & 62.03 \\
        15-31 & Sparse & Linear & 52.90 & 83.1 & 59.6 & 62.12 \\
        (0-15)-(16-30)-31 & Dense & DPT & 54.27 & 85.4 & 60.7 & 63.65 \\
        15-31 & Sparse & DPT & 54.58 & 84.6 & 61.0 & 63.70 \\            
        7-15-23-31 & Sparse & DPT & \B{55.19} & 85.9 & 61.6 & 64.46 \\        
        (0-9)-(10-19)-(20-30)-31 & Dense & DPT & 54.28 & 85.5 & 62.3 & 64.08 \\ 
        3-7-11-15-19-23-27-31 & Sparse & DPT & 54.42 & \B{86.7}  & \B{62.8} & \textbf{64.58} \\
    \midrule
      &  & \B{TextVQA} & \B{ChartQA} & \B{DocVQA} &  \B{InfoVQA} & \B{OCRBench} \\
        \midrule        
        Last & N/A & \B{63.6} & 23.4 & 47.0 & 33.8 & \B{42.0} \\
        7-15-23-31 & Sparse & 63.2 & \B{24.1} & 
        \B{47.2} & \B{34.3} & 40.3 \\
        (0-9)-(10-19)-(20-30)-31 & Dense & 63.5 & 23.1 & 47.0 & 33.5 & 40.2 \\
    \bottomrule
    \end{tabular}
    }    
    
    \caption{Study on the effect of RADIOv2.5-H feature selection. For ``dense'' aggregation, the numbers in brackets indicate the range of layers from which the average is calculated. \B{Top:} Semantic segmentation, depth estimation, surface normals estimation. \B{Bottom}: VILA benchmarks. Pixel-level tasks exhibit a clear preference for more layers and non-linear heads, while VLM tasks seem mostly neutral to this choice.}
    \label{tab:dense-features-semseg}
\end{table}

\section{Experiments}

\begin{table*}[ht]
    \centering
    \resizebox{\textwidth}{!}{
    \begin{tabular}{l|c|c|ccc|cccc}
        \toprule
        \multirow{3}{*}{\B{Model}} & \multirow{3}{*}{\B{Params}} & \B{ADE20k} & \multicolumn{3}{c|}{\B{NYUd}} & \multicolumn{4}{c}{\B{Pascal Context}} \\
        &  & SemSeg & SemSeg & Depth & Normal & SemSeg & Parsing & Saliency & Normal \\
        
        &  &  mIoU$\uparrow$ &  mIoU$\uparrow$ &  RMSE$\downarrow$ &  mErr$\downarrow$ &  mIoU$\uparrow$ &  mIoU$\uparrow$ & maxF$\uparrow$ &  mErr$\downarrow$ \\
        \midrule
        SAM-CLIP\cite{wang2024samclipmergingvisionfoundation} & 86M & 38.4 & †  & †  & †  & †  & †  & †  & †  \\
        Theia~\cite{shang2024theiadistillingdiversevision} & 86M & 35.55 & 38.90      & 0.6377      & 24.11  & 69.84      & 60.67       & 80.63      & 16.94 \\
        UNIC-B\cite{sariyildiz2024unicuniversalclassificationmodels} & 86M & \underline{39.6} & \underline{42.21}      & 
        
    \underline{0.6172}      & \underline{22.78} & \underline{75.90}      & \underline{62.85}       & \B{81.84} & \B{15.78} \\
        RADIOv2.5-B (ours) & 98M & \B{48.94} & \B{57.19} & \B{0.4980} & \B{20.04} & \B{81.75} & \B{71.49}       & \underline{81.26}      & \underline{16.10}   \\
        \midrule
        UNIC-L\cite{sariyildiz2024unicuniversalclassificationmodels} & 307M & \underline{48.30} & \underline{58.56}      & \underline{0.4916}      & \underline{19.34}  & \underline{81.82}      & \underline{72.24}       & \underline{79.21}      & \underline{17.35}  \\        
        RADIOv2.5-L (ours) & 320M & \B{52.95} & \B{61.42}      & \B{0.4577}      & \B{18.57} & \B{82.87} & \B{74.32}       & \B{81.65}      & \B{16.15} \\        
        \midrule
        UNIT\cite{zhu2024unitunifyingimagetext} & 632M & 50.19 & 61.44 & 0.4538 & 19.07 & 82.36 & 73.34 & \underline{78.70} & 17.98  \\
        RADIOv2.1-H & 653M & \underline{51.36} & \underline{62.76}      & \B{0.4339} & \underline{18.43}  & \underline{82.78}      & \underline{74.42}       & 78.48      & \underline{17.53}  \\
        RADIOv2.5-H & 653M & \B{53.97} & \B{63.82} & \underline{0.4353}      & \B{17.67} & \B{83.43} & \B{75.75}  & \B{81.19}      & \B{16.16}  \\  
        \midrule
        UNIT\cite{zhu2024unitunifyingimagetext} & 1B & \underline{50.63} & \underline{61.62} & \underline{0.4527} & \underline{19.07} & \underline{82.20} & \underline{73.22} & \underline{78.51} & \underline{17.96} \\
        RADIOv2.5-g (ours) & 1.1B & \B{54.56} &  \B{63.46}      & \B{0.4318}      & \B{17.08} & \B{82.65}      & \B{75.18}       & \B{80.08}      & \B{17.06} \\
    \end{tabular}
    }
    \caption{Comparison of dense task benchmarks against other agglomerative models. † Results unavailable due to lack of access to SAM-CLIP model weights. \textit{Note: RADIO requires more parameters for a given model class due to a $C \times 128 \times 128$ position embedding buffer coming from CPE, but does not contribute meaningfully to increased FLOPs or representation capacity.} }
    \label{tab:related-agglomerative-models-summary}
\end{table*}

\begin{table*}[ht]
\centering\resizebox{\textwidth}{!}{
    \begin{tabular}{lccccccccccccccc}
        \toprule
\textbf{Vision Encoder} & \rotatebox{90}{\B{Resolution}} & \rotatebox{90}{\B{Tokens/im}} &  \multicolumn{1}{|c}{\rotatebox{90}{\B{TextVQA}}}  & \rotatebox{90}{\B{ChartQA}} & \rotatebox{90}{\B{DocVQA}} & \rotatebox{90}{\B{InfoVQA}} & \rotatebox{90}{\B{OCRBench}} & \rotatebox{90}{\B{GQA}} & \rotatebox{90}{\B{POPE}} & \rotatebox{90}{\B{MME}} & \rotatebox{90}{\B{SEED(I)}} 
& \rotatebox{90}{\B{MMMU}} &\multicolumn{1}{c|}{\rotatebox{90}{\B{AI2D}}} & \rotatebox{90}{\B{Average}} \\
\midrule
OpenAI-CLIP~\cite{radford2021learningtransferablevisualmodels} & $336^2$ & 196 & \multicolumn{1}{|c}{63.2} & 49.2 & 43.4 & 27.4 & 431 & 62.4 & 85.7 & 1517.9 & 72.7 & 43.4 & 77.6 & \multicolumn{1}{|c}{58.55} \\
RADIOv2.1-H~\cite{ranzinger2024radio} & $512^2$ & 196  & \multicolumn{1}{|c}{63.6} &  51.7 & 40.1 & 26.2 & 392 & \underline{63.4} & 87.3 & 1570.6 & 73.6 & \B{43.9} & 78.9 & \multicolumn{1}{|c}{58.76} \\
RADIOv2.1-H~\cite{ranzinger2024radio} & $768^2$ & 196  & \multicolumn{1}{|c}{62.8} & 49.9 & 36.7 & 24.8 & 365 & 62.7 & 87.0 & 1517.2 & 71.7 & 41.9& 73.4 & \multicolumn{1}{|c}{56.66} \\
SigLIP-SO400M~\cite{zhai2023sigmoidlosslanguageimage} & $384^2$ & 196 & \multicolumn{1}{|c}{64.8} & 55.7 & 47.9 & 29.2 & 452 & 63.1 & 85.4 & \B{1632.9} & 72.9 & 41.1&80.7 & \multicolumn{1}{|c}{60.95} \\
\midrule
RADIOv2.5-L (ours) & $768^2$ & 196 & \multicolumn{1}{|c}{66.7} & 56.4 & 49.2 & 29.8 & 441 & \underline{63.4} & \B{87.6} & 1592.4 & 74.0 & 43.3 & 79.2 & \multicolumn{1}{|c}{61.21} \\
RADIOv2.5-H (ours) & $768^2$ & 196 & \multicolumn{1}{|c}{\underline{68.8}} & \B{61.6} & \B{54.2} & \underline{30.8} & \B{498} & \B{63.8} & 87.3 & 1562.3 & \underline{74.1} & \B{44.1}& \B{82.5} & \multicolumn{1}{|c}{\B{63.19}} \\
RADIOv2.5-g (ours) & $672^2$ & 196 & \multicolumn{1}{|c}{\B{69.6}} & \underline{57.0} & \underline{53.3} & \B{30.9} & \underline{479} & \underline{63.4} & \underline{87.5} & \underline{1606.4} & \B{74.3} & 41.3& \underline{80.8} & \multicolumn{1}{|c}{\underline{62.39}} \\
        \bottomrule
    \end{tabular}
    }    
        \caption{Vision Encoder Comparison in VILA\cite{lin2023vila}. We used \B{MN-Minitron-8B}\cite{sreenivas2024llmpruningdistillationpractice} as LLM and the LLaVA1.6\cite{liu2024llavanext} data mixture. We use Token Merging to compress all visual features down to 196 tokens. From left to right we report: image resolution, number of tokens per image, TextVQA (with OCR hints) validation accuracy, ChartQA overall, DocVQA validation accuracy, InfoVQA validation accuracy, OCRBench accuracy, GQA (TestDev) accuracy, POPE F1 score, MME perception score, SEED (Image) accuracy, MMMU (validation) accuracy, AI2D (no mask) accuracy, average (calculated after dividing MME score by 20 and OCR score by 10).}
    \label{tab:vlm-vision-encoder-comparison-summary}
    \vspace{-4mm}
\end{table*}

\subsection{Evaluation Benchmarks}

For Image classification, we report the ImageNet-1k\cite{russakovsky2015imagenetlargescalevisual} Top-1 classification accuracy using two methods. \textit{i)} zero-shot\cite{radford2021learningtransferablevisualmodels} classification using the DFN CLIP language model; \textit{ii)} k-NN\cite{caron2021emergingpropertiesselfsupervisedvision} classification using the embeddings of the shared backbone.

For Instance Segmentation, we use the EfficientViT\cite{liu2023efficientvitmemoryefficientvision} implementation of a COCO\cite{kim2023regionawarepretrainingopenvocabularyobject} instance segmenter to measure how well our SAM adaptor head can replace SAM.

For Semantic Segmentation, we use the MMSeg\cite{mmseg2020} framework, and train a linear probe on top of the frozen RADIOv2.5 features. We use the ADE20k\cite{zhou2018semanticunderstandingscenesade20k} dataset. We train the probe at three resolutions ($512^2$, $768^2$, and $1024^2$) and report the best result (full results in appendix).

For 3D understanding experiments we use the Probe3D\cite{elbanani2024probing} code base and apply a DPT head on top of the frozen backbone features. We train and evaluate on the Navi\cite{jampani2023navi} dataset. We resize images to $512^2$ during both training and evaluation.

For Vision-Language Modeling we follow the same framework as in VILA~\cite{lin2023vila}. We pair our vision encoder with a MN-Minitron-8B\cite{sreenivas2024llmpruningdistillationpractice} LLM. We train all alignment/pretraining/SFT and unfreeze the vision encoder during SFT. We train the model using ShareGPT4v\cite{chen2023sharegpt4v} and VFLAN\cite{xu2024vision} datasets.
We leverage insights from Cambrian\cite{tong2024cambrian1fullyopenvisioncentric} to select the benchmarks that are most sensitive to the quality of visual features  report scores on TextVQA\cite{singh2019vqamodelsread}, ChartQA\cite{masry2022chartqabenchmarkquestionanswering}, DocVQA\cite{mathew2021docvqadatasetvqadocument}, InfoVQA\cite{mathew2021infographicvqa} and OCRBench\cite{liu2024ocrbenchhiddenmysteryocr}.

\subsection{Ablation Studies}

Ablation studies are summarized in Table \ref{tab:ablation_table}. Our initial RADIOv2.1-L baseline (config \B{$\mathcal{A}$}) is a reproduction of AM-RADIO~\cite{ranzinger2024radio}, using a ViT-L backbone and the DataComp1B\cite{gadre2023datacompsearchgenerationmultimodal} dataset. 
Our hyperparameters are detailed in the appendix.

We add multi-resolution training in config \B{$\mathcal{B}$}, and notice improved results on most dense tasks (+2.8 on ADE20k, +3.4 on average on VLMs). Multi-score robustness if particularly evident on Figure~\ref{fig:zero-shot-accuracy-vs-resolution}, where config \B{$\mathcal{B}$} exhibits much more consistent results at higher resolutions.

In config \B{$\mathcal{C}$}, we replace the OpenAI CLIP teacher with SigLIP~\cite{zhai2023sigmoidlosslanguageimage} and observe a 6.7-point improvement on VLM benchmarks.

In config \B{$\mathcal{D}$}, we replace the ViT-L 320M-parameter backbone with a 553M-parameter ViT-H backbone, which improves all metrics significantly.

Config \B{$\mathcal{E}$} applies only to VLMs, where we replace $2^2$ pixel unshuffling with Token Merging. This reduces the number of vision tokens from 576 to 256 and improves VLM scores by an average of 5.9 points. Appendix Table~\ref{tab:effect-of-compression-on-siglip-and-radio} provides a detailed analysis of the trade-off between the number of vision tokens and accuracy, highlighting RADIOv2.5's comparative advantage over SigLIP in terms of token merging effectiveness.

\subsection{Comparative Results}

\subsubsection{Dense Task Evaluation}

We evaluate our models in comparison to other agglomerative models of similar size (SAM-CLIP~\cite{wang2024samclipmergingvisionfoundation}, Theia~\cite{shang2024theiadistillingdiversevision}, UNIC~\cite{sariyildiz2024unicuniversalclassificationmodels}, UNIT~\cite{zhu2024unitunifyingimagetext}). Following MLoRE~\cite{jiang2024mlore} we report metrics on NYUDv2 and PASCAL Context for our model. We train with a learning rate of $1e-3$, and use a weight of $1$ for all tasks. We purposefully don't tune any hyperparameters. We keep the backbone model frozen, and use MLoRE's ``conv'' head for each task. The conv head is defined as follows: 
\(\textup{Conv-3x3} \rightarrow \textup{BatchNorm} \rightarrow \textup{GeLU} \rightarrow \textup{Conv-1x1}\).

Results are reported in Table~\ref{tab:related-agglomerative-models-summary}. RADIOv2.5 models exhibit a consistent improvement over previous baselines.

\subsubsection{VLM Evaluation}

We compare our models against state-of-the-art vision encoders: OpenAI-CLIP~\cite{radford2021learningtransferablevisualmodels}, RADIOv2.1-H and SigLIP-SO400M~\cite{zhai2023sigmoidlosslanguageimage}. We pair the vision encoders with MN-Minitron-8B~\cite{sreenivas2024llmpruningdistillationpractice} and train them in VILA~\cite{lin2023vila} using the LLaVA1.6~\cite{liu2024llavanext} data mixture. For all vision encoders, we use Token Merging to compress the visual features to 196 tokens in order to ensure a fixed cost to the Language Model. We report results in Table~\ref{tab:vlm-vision-encoder-comparison-summary}. We observe that RADIOv2.5 models consistently exceed the accuracy of other vision encoders.
\section{Related Work}

\subsection{Agglomerative Models}

AM-RADIO~\cite{ranzinger2024radio} and SAM-CLIP~\cite{wang2024samclipmergingvisionfoundation} have pioneered multi-teacher distillation for Vision Foundation Models. However, AM-RADIO exhibits inconsistent accuracy across resolutions, while SAM-CLIP's lack of a well-performing dense model, such as DINO, leads to low accuracy in semantic segmentation tasks.
Theia~\cite{shang2024theiadistillingdiversevision} employs a multi-distillation framework for robotic applications, disregarding summary tokens and selecting a different set of teachers, including DepthAnything~\cite{Yang2024DepthAU}, to demonstrate how each contributes to robot performance.
UNIC~\cite{sariyildiz2024unicuniversalclassificationmodels} stabilizes training with a ladder of expandable projectors and preserves the accuracy of top teachers through teacher-dropping regularization. Our method differs by using teacher feature standardization to balance teachers while focusing on maintaining accuracy across scales, through multi-resolution training.

\subsection{Multi-Resolution Support}

The original ViT~\cite{dosovitskiy2021imageworth16x16words} suggests interpolating pre-trained position embeddings to enable fine-tuning at different scales. FlexiViT~\cite{Beyer2022FlexiViTOM} trains Vision Transformers at multiple patch sizes, allowing flexible adaptation to various computational budgets without retraining. NaViT~\cite{Dehghani2023PatchNP} processes images of varying resolutions and aspect ratios by employing sequence packing during training. Our method differs  in two key ways: by leveraging Cropped Position Embeddings (CPE)~\cite{kim2023regionawarepretrainingopenvocabularyobject} for robust inference across scales and also by devising strategies to distill a resolution-robust student from inflexible teachers of varying resolutions.

\subsection{Multi-Encoder VLMs}

Cambrian-1~\cite{tong2024cambrian1fullyopenvisioncentric} integrates multiple vision encoders through a Spatial Vision Aggregator, utilizing learnable latent queries that interact with several layers of a Large Language Model (LLM) via cross-attention.
Eagle~\cite{shi2024eagleexploringdesignspace} combines multiple vision experts by concatenating their visual tokens along the channel dimension.
BRAVE~\cite{kar2024brave} introduces the Multi-Encoder Querying Transformer, which merges visual features from various encoders into a fixed-length representation, subsequently used as a soft prompt for a frozen language model. In contrast, our method performs multi-vision encoder aggregation as a preliminary step to develop a single, versatile vision encoder suitable for integration into a VLM.

\subsection{Applications of Agglomerative Models}

Agglomerative models are seeing widespread adoption in Computer Vision tasks. In \cite{wang2025skilsemantickeypointimitation}, RADIOv2.5-B demonstrates an excellent efficiency tradeoff, as its robust features are utilized for Semantic Keypoint tracking in robotic hand manipulation tasks. In \cite{zhou2025sameexploringvisualcorrespondence}, RADIO features are used alongside those of \cite{chen2024far} within a VLM to enhance multi-image correspondence.

Believing that VLMs can become generalists in Computer Vision, we dedicate significant effort to their study.

\section{Conclusion}

In this paper, we presented RADIOv2.5, a robust framework for multi-teacher vision model distillation that accommodates variations in teacher resolutions, activation magnitudes, and noise patterns.
The key differentiating factor in our work is our focus on preserving accuracy across a broad range of resolutions.
Our multi-resolution training approach mitigates resolution-dependent performance degradation (mode switching).
Our findings show that mosaic augmentation and PHI-S effectively balance computational load and loss contributions from each teacher.

Token compression enables efficient integration with VLMs by preserving critical visual information, even at high compression ratios. Our feature selection study highlighted the advantages of using intermediate activations for dense tasks, particularly when non-linear transformations are used.

{
    \small
    \bibliographystyle{ieeenat_fullname}
    \bibliography{main}
}

\appendix
\renewcommand{\thesection}{A.\arabic{section}}
\renewcommand{\thetable}{A\arabic{table}}
\renewcommand{\thefigure}{A\arabic{figure}}
\setcounter{section}{0}
\setcounter{table}{0}
\setcounter{figure}{0}

\clearpage

\clearpage

\setcounter{page}{1}
\maketitlesupplementary

\onecolumn

\section{VLM Benchmarks}

\subsection{More Vision Encoder Comparisons}

\begin{table}[h]
\centering\resizebox{\linewidth}{!}{
    \begin{tabular}{lcccccccccccccc}
        \toprule
\textbf{Vision Encoder} & \rotatebox{90}{\B{Resolution}} & \rotatebox{90}{\B{Tokens/im}} &  \multicolumn{1}{|c}{\rotatebox{90}{\B{TextVQA}}}  & \rotatebox{90}{\B{ChartQA}} & \rotatebox{90}{\B{DocVQA}} & \rotatebox{90}{\B{InfoVQA}} & \rotatebox{90}{\B{OCRBench}} & \rotatebox{90}{\B{GQA}} & \rotatebox{90}{\B{POPE}} & \rotatebox{90}{\B{MME}} & \rotatebox{90}{\B{SEED(I)}}  &\multicolumn{1}{c|}{\rotatebox{90}{\B{AI2D}}} & \B{Average} \\
\midrule
OpenAI-CLIP~\cite{radford2021learningtransferablevisualmodels} & $336^2$ & 144 & \multicolumn{1}{|c}{63.8} & 27.5  & 48.8 & 33.0 & 414 & 63.0 & 86.7 & 1646.9 & 66.7 & 67.1 & \multicolumn{1}{|c}{58.03} \\
AM-RADIO-H~\cite{ranzinger2024radio} & $512^2$ & 256 & \multicolumn{1}{|c}{55.9} &  15.7  & 35.2  & 30.6  & 316  & 60.2  & 85.3 & 1516.5  & 71.8 & 63.1  & \multicolumn{1}{|c}{52.52} \\
SigLIP-SO400M~\cite{zhai2023sigmoidlosslanguageimage} & $384^2$ & 196 & \multicolumn{1}{|c}{67.6} & 33.0 & 57.1 & 36.0 & 458 & 63.0 & 85.7 & 1605.2 & 74.2 & \B{68.6} & \multicolumn{1}{|c}{61.13} \\

\B{RADIOv2.5-H (ours)} & $768^2$ & 196 & \multicolumn{1}{|c}{\B{70.8}} &  33.9 & 59.5 & 37.0 & 482 & 63.9 & 86.9 & \B{1613.5} & 75.1 & 66.7 & \multicolumn{1}{|c}{62.27} \\
\B{RADIOv2.5-H (ours)} & $768^2$ & 512 & \multicolumn{1}{|c}{70.2} &  \B{37.3} & \B{64.2} & \B{37.6} & \B{523} & \B{64.5} & \B{87.3} & 1587.6 & \B{75.4} & 67.5 & \multicolumn{1}{|c}{\B{63.57}} \\
        \bottomrule
    \end{tabular}
    }
    
        \caption{VILA benchmark results for various vision encoders. We used \B{Qwen2-7B-Instruct} as LLM and \B{ShareGPT4v\cite{chen2023sharegpt4v} and VFLAN\cite{xu2024vision}} data. From left to right we report: image resolution, numbers of tokens per image, TextVQA (with OCR hints) validation accuracy, ChartQA overall, DocVQA validation accuracy, InfoVQA validation accuracy, OCRBench accuracy, GQA (TestDev) accuracy, POPE F1 score, MME perception score, SEED (Image) accuracy, AI2D accuracy, average (calculated after dividing MME score by 20 and OCR score by 10).}
    \label{tab:vlm-vision-encoder-comparison}
    \vspace{-4mm}
\end{table}

In Table~\ref{tab:vlm-vision-encoder-comparison}, we report benchmark results for OpenAI-CLIP-336, AM-RADIO-H, SigLIP-400m, and RADIOv2.5-H. The same Qwen2-7B-Instruct LLM, training data, and hyperparameters are used across all configurations. RADIOv2.5-H is utilized in conjunction with Token Merging and configured to produce either 196 tokens per image (matching SigLIP) or 512 tokens per image (to demonstrate scaling capabilities). The results show a significant improvement when transitioning to RADIOv2.5-H, even with the same token count as the SigLIP baseline. Furthermore, additional benefits are observed when scaling up the number of vision tokens.

\subsection{Scaling up to More Data}

\begin{table}[h]
\centering\resizebox{\linewidth}{!}{
    \begin{tabular}{lcccccccccccc}
        \toprule
\textbf{Vision Encoder} &  \multicolumn{1}{|c}{\B{TextVQA}}  & \B{ChartQA} & \B{DocVQA} & \B{InfoVQA} & \B{OCRBench} & \B{GQA} & \B{POPE} & \B{MME} & \B{SEED(I)}  &\multicolumn{1}{c|}{\B{AI2D}} & \B{Average} \\
\midrule
SigLIP-SO400M~\cite{zhai2023sigmoidlosslanguageimage} & \multicolumn{1}{|c}{69.7} & 67.2 & 63.7 & 40.9 & 588 & 62.4 & 86.9 & 1648.5 & 72.0 & \B{75.0} & \multicolumn{1}{|c}{67.90} \\
\B{RADIOv2.5-H (ours)} & \multicolumn{1}{|c}{\B{71.5}} & \B{71.9} & \B{73.6} & \B{45.6} & \B{667} & \B{62.7} & \B{87.5} & \B{1664.1} & \B{77.39} & 74.8 & \multicolumn{1}{|c}{\B{71.49}} \\
        \bottomrule
    \end{tabular}
    }
    
        \caption{VILA benchmark results for SigLIP and RADIOv2.5-H. We used \B{Qwen2-7B-Instruct~\cite{wang2024qwen2vlenhancingvisionlanguagemodels}} as LLM, with an improved data mixture of public datasets. From left to right we report: TextVQA (with OCR hints) validation accuracy, ChartQA overall, DocVQA validation accuracy, InfoVQA validation accuracy, OCRBench accuracy, GQA (TestDev) accuracy, POPE F1 score, MME perception score, SEED (Image) accuracy, AI2D accuracy, average (calculated after dividing MME score by 20 and OCR score by 10).}
    \label{tab:vlm-scaling-to-more-data}
    \vspace{-4mm}
\end{table}

In Table~\ref{tab:vlm-scaling-to-more-data}, we report benchmark results obtained using an improved SFT data mixture, which includes data from ShareGPT4v, LLaVA Instruct, Cambrian, VFLAN, and the training sets of some benchmarks, for a total of 9.8M samples. RADIOv2.5-H is used in conjunction with Token Merging and is configured to produce 512 tokens per image. RADIOv2.5-H outperforms the SigLIP baseline on all benchmarks, except for AI2D, where it achieves a tie with SigLIP.

\subsection{Effect of the Compression Method}

\begin{table}[h]
\centering\resizebox{\linewidth}{!}{
    \begin{tabular}{lcccccccccccccc}
        \toprule
\textbf{Vision Encoder} & \textbf{Compression}  & \rotatebox{90}{\textbf{Tokens/im}} &  \multicolumn{1}{|c}{\rotatebox{90}{\B{TextVQA}}}  & \rotatebox{90}{\B{ChartQA}} & \rotatebox{90}{\B{DocVQA}} & \rotatebox{90}{\B{InfoVQA}} & \rotatebox{90}{\B{OCRBench}} & \rotatebox{90}{\B{GQA}} & \rotatebox{90}{\B{POPE}} & \rotatebox{90}{\B{MME}} & \rotatebox{90}{\B{SEED(I)}}  &\multicolumn{1}{c|}{\rotatebox{90}{\B{AI2D}}} & \B{Average} \\
\midrule
SigLIP SO400M~\cite{zhai2023sigmoidlosslanguageimage} & $2 \times 2$ Unshuffle & \multicolumn{1}{c|}{196} & \B{63.4} & \B{23.1} & 41.4 & \B{30.1} & \B{339} & 63.8 & 85.4 & 1518.4 & 70.9 & 63.6 & \multicolumn{1}{|c}{\B{55.15}} \\
SigLIP SO400M~\cite{zhai2023sigmoidlosslanguageimage} & ToMe r=533 & \multicolumn{1}{c|}{196} & 62.9 & 21.2 & \B{41.8} & \B{30.1} & 326 & \B{64.2} & \B{85.8} & \B{1537.5} & \B{71.7}  &\B{63.7} & \multicolumn{1}{|c}{55.09} \\        
        \midrule
\B{RADIOv2.5-H (ours)} & $2 \times 2$ Unshuffle & \multicolumn{1}{c|}{576} & 66.4 & 25.0 & 53.3 & 32.5 & 402 & \B{64.8} & 86.5 & 1434.0 & 73.4 & 63.9 & \multicolumn{1}{|c}{57.77 } \\        
\B{RADIOv2.5-H (ours)} & ToMe r=2048 & \multicolumn{1}{c|}{256} & \B{69.7} & 30.4 & 52.3 & 36.2 & 429 & 63.8 & 86.8 & \B{1572.4} & \B{74.6}  & 65.1 & \multicolumn{1}{|c}{60.04} \\ 
\B{RADIOv2.5-H (ours)} & ToMe r=3584 & \multicolumn{1}{c|}{512} & 68.9 & \B{31.3} & \B{53.9} & \B{37.0} & \B{446} & 63.0 & \B{87.6} & 1537.7  & 73.9 & \B{66.0} & \multicolumn{1}{|c}{\B{60.31}} \\   
        \bottomrule
    \end{tabular}
    }
    
        \caption{VILA benchmark results for various vision encoders and compression methods. We used \B{MN-Minitron-8B} as LLM and \B{ShareGPT4v\cite{chen2023sharegpt4v} and VFLAN\cite{xu2024vision}} data. From left to right we report: number of vision tokens per image, TextVQA (with OCR hints) validation accuracy, ChartQA overall, DocVQA validation accuracy, InfoVQA validation accuracy, OCRBench accuracy, GQA (TestDev) accuracy, POPE F1 score, MME perception score, SEED (Image) accuracy, AI2D accuracy, average (calculated after dividing MME score by 20 and OCR score by 10).}
    \label{tab:effect-of-compression-on-siglip-and-radio}
    \vspace{-4mm}
\end{table}

In Table~\ref{tab:effect-of-compression-on-siglip-and-radio}, we report benchmark results for SigLIP-400m and RADIOv2.5-H using different token compression methods. The same MN-Minitron-8B LLM, training data, and hyperparameters are used across all configurations. The results show no improvement when applying Token Merging to SigLIP. RADIOv2.5-H, on the other hand, demonstrates significant improvement with Token Merging, and increasing the token count from 256 to 512 provides a modest additional gain.

\subsection{Block-wise vs Output Token Merging}

\begin{table*}[ht]
\centering\resizebox{\textwidth}{!}{
    \begin{tabular}{lccccccccccccccc}
        \toprule
\textbf{Token Merging} & \rotatebox{90}{\B{Resolution}} & \rotatebox{90}{\B{Tokens/im}} &  \multicolumn{1}{|c}{\rotatebox{90}{\B{TextVQA}}}  & \rotatebox{90}{\B{ChartQA}} & \rotatebox{90}{\B{DocVQA}} & \rotatebox{90}{\B{InfoVQA}} & \rotatebox{90}{\B{OCRBench}} & \rotatebox{90}{\B{GQA}} & \rotatebox{90}{\B{POPE}} & \rotatebox{90}{\B{MME}} & \rotatebox{90}{\B{SEED(I)}} 
& \rotatebox{90}{\B{MMMU}} &\multicolumn{1}{c|}{\rotatebox{90}{\B{AI2D}}} & \rotatebox{90}{\B{Average}} \\
\midrule
Block-wise ($r=\sim 65$) & $768^2$ & 196 & \multicolumn{1}{|c}{65.2} & 61.3 & 52.4 & 29.3 & 469 & 63.4 & \B{87.7} & 1539.4 & 73.9 & 42.8 & 79.5 & \multicolumn{1}{|c}{61.76} \\
Output ($r=2108$) & $768^2$ & 196 & \multicolumn{1}{|c}{\B{68.8}} & \B{61.6} & \B{54.2} & \B{30.8} & \B{498} & \B{63.8} & 87.3 & \B{1562.3} & \B{74.1} & \B{44.1}& \B{82.5} & \multicolumn{1}{|c}{\B{63.19}} \\
        \bottomrule
    \end{tabular}
    }    
        \caption{Comparison of block-wise vs output token merging. We use a RADIOv2.5-H vision encoder.
        \B{``Block-wise'' token merging}: we merge tokens in each successive ViT block, using keys as criteria. We assign 50\% of tokens as target tokens and set $r=65$ for the first 31 blocks and $r=93$ for the last block, totaling 2108 merged tokens and bringing the final number of tokens to 196.
        \B{``Output'' token merging}: we only merge the ViT output tokens, using token values as criteria. We partition source/targets tokens in a $ 6 \times 6 $ strided pattern and set r=2108.}
    \label{tab:token-merging-inline-vs-output}
    \vspace{-4mm}
\end{table*}

In this section we evaluate which works better: merging tokens incrementally in each ViT block (using ``keys'' as criteria as in the original ToMe~\cite{bolya2023tokenmergingvitfaster} formulation), or merging tokens once at the output of the ViT. Results are shown in Table~\ref{tab:token-merging-inline-vs-output} and indicate that in our setup, one-shot merging at the output yields improved results.


\subsection{High-Resolution Inference using Tiling}

\begin{table}[h]
\centering\resizebox{\linewidth}{!}{
    \begin{tabular}{lcccccccccccccccc}
        \toprule
 \textbf{Vision Encoder} & \textbf{Resolution} & \textbf{Compression}  & \rotatebox{90}{\textbf{Tokens/im}} &  \multicolumn{1}{|c}{\rotatebox{90}{\B{TextVQA}}}  & \rotatebox{90}{\B{ChartQA}} & \rotatebox{90}{\B{DocVQA}} & \rotatebox{90}{\B{InfoVQA}} & \rotatebox{90}{\B{OCRBench}} & \rotatebox{90}{\B{GQA}} & \rotatebox{90}{\B{POPE}} & \rotatebox{90}{\B{MME}} & \rotatebox{90}{\B{SEED(I)}}  &\multicolumn{1}{c|}{\rotatebox{90}{\B{AI2D}}} & \B{Average} \\
\midrule
SigLIP SO400M~\cite{zhai2023sigmoidlosslanguageimage} + Tiling & Up to $13 \times 384^2 $ & $2 \times 2$ Unshuffle & \multicolumn{1}{c|}{$ \sim 1928$} & 66.5 & \B{64.6} & \B{74.4} & 40.1 & 521 & \B{64.5} & \B{88.0} & 1536.1 & 74.4 & 79.3 & \multicolumn{1}{|c}{68.07} \\
\B{RADIOv2.5-H (ours)} & $768^2$ & ToMe r=2108 & \multicolumn{1}{c|}{196} & 68.8 & 61.4 & 61.9 & 36.7 & 498 & 63.8 & 88.4 & 1562.6 & 74.1 & 71.9  & \multicolumn{1}{|c}{65.49} \\ 
\B{RADIOv2.5-H (ours) + Tiling} & Up to $7 \times 768^2$ & ToMe r=2108 & \multicolumn{1}{c|}{$ \sim 1233$} & \B{73.2} & 63.5 & 73.1 & \B{46.5} & \B{545} & \B{64.5} & 87.9 & \B{1611.6} & \B{75.3} & \B{82.4}  & \multicolumn{1}{|c}{\B{70.15}} \\ 
\bottomrule
    \end{tabular}
    }
    
        \caption{VILA benchmark results using tiling to emulate high-resolution inference. We used \B{MN-Minitron-8B} as LLM and the data mixture from \B{LLaVA1.6}. From left to right we report: number of vision tokens per image (calculated across all benchmarks), TextVQA (with OCR hints) validation accuracy, ChartQA overall, DocVQA validation accuracy, InfoVQA validation accuracy, OCRBench accuracy, GQA (TestDev) accuracy, POPE F1 score, MME perception score, SEED (Image) accuracy, AI2D (No Mask) accuracy, average (calculated after dividing MME score by 20 and OCR score by 10).}
    \label{tab:tiling-vs-high-res}
    \vspace{-4mm}
\end{table}

In Table~\ref{tab:tiling-vs-high-res}, we emulate high-resolution inference using tiling~\cite{chen2024far}. While we acknowledge that there is no strict equivalence in the number of pixels (SigLIP processes an average of 1.8M pixels per sample, whereas RADIOv2.5-H processes an average of 3.9M pixels) or in the number of generated tokens (SigLIP outputs an average of 2,410 tokens per sample, compared to 1,313 tokens for RADIOv2.5), we observe slightly improved accuracy with RADIOv2.5 despite producing a significantly smaller number of output tokens.

\section{Semantic Segmentation}

\begin{table}[h]
\centering
    \begin{tabular}{lcccccccc}
        \toprule
        \multirow{2}{*}{\textbf{Model}} &  \multirow{2}{*}{\B{Params}} & \multicolumn{3}{|c}{\B{ADE20k}} & \multicolumn{3}{|c}{\B{Pascal VOC}} \\ 
                                &        & \multicolumn{1}{|c}{512} & 768 & 1024  
                                        & \multicolumn{1}{|c}{512} & 768 & 1024 \\                                        
        \midrule        
        \B{RADIOv2.5-B (ours)} & 98M & \B{48.94} & \B{50.48} & \B{51.16} & \B{84.35}  & \B{85.47} & \B{85.33} \\  
        \midrule
        AM-RADIO-L~\cite{ranzinger2024radio}*   & 320M & 50.03 & 37.99 & 35.63 & 83.76 & 68.85 & 63.86 & \\
        + multi-res & 320M & \B{51.54} & 51.74 & 52.84 & \B{85.51} & 86.84 & \B{87.21} \\
        \B{RADIOv2.5-L (ours)} & 320M & 51.47 & \B{51.90} & \B{52.95} &  85.49 & \B{86.96} & 87.03 \\        
        \midrule         
        AM-RADIO-H~\cite{ranzinger2024radio}*   & 553M & 51.34 & 35.78 & 32.99 &  84.71 & 64.54 & 59.15 & \\
        \B{RADIOv2.5-H (ours)}  & 553M & \B{51.58} & \B{52.45} &  \B{53.91}  & \B{85.97} & \B{87.54} & \B{87.69} \\
        \midrule
        DINOv2-g-reg   & 1.1B & 48.79 & 48.37 & 50.71 & 82.72 & 83.95 & 84.25 \\
        \bottomrule

    \end{tabular}

        \caption{Semantic segmentation mIoU for ADE20k and Pascal VOC across different models and resolutions: $512 \times 512$, $768 \times 768$, and $1024 \times 1024$. Note: For DINOv2, we use the nearest larger multiple of its patch size (14). We apply a linear probe on top of the frozen features from the vision encoder. The mIoU of RADIOv2.5-B exceeds that of DINOv2-g-reg, despite being only one-tenth the size. *Our reproduction.}
    \label{tab:semseg-resolutions}

\end{table}

In Table~\ref{tab:semseg-resolutions}, we report Semantic Segmentation mIoU at different resolutions on ADE20k~\cite{zhou2018semanticunderstandingscenesade20k} and VOC~\cite{Everingham15}. We observe that RADIOv2.5 favorably scales to higher resolutions, while the accuracy of AM-RADIO falls above a resolution of $512 \times 512$.

\section{Additional Benchmarks}

Following MLoRE~\cite{jiang2024mlore} we report metrics on NYUDv2 and PASCAL Context for our model, along with the other agglomerative models (Theia~\cite{shang2024theiadistillingdiversevision} and UNIC~\cite{sariyildiz2024unicuniversalclassificationmodels}), as well as DINOv2~\cite{darcet2024visiontransformersneedregisters} as it's the core perception teacher for these types of tasks. Previously to this study, MLoRE was the state-of-the-art for producing a multi-task model for these metrics. We keep the backbone model frozen, and use MLoRE's ``conv'' head for each task. The conv head is defined as follows:

\begin{equation}
    \textup{Conv-3x3} \rightarrow \textup{BatchNorm} \rightarrow \textup{GeLU} \rightarrow \textup{Conv-1x1}
\end{equation}

At the largest scale, RADIOv2.5-H is extremely competitive with DINOv2-g at half the parameters. At the ViT-B and ViT-L scale, RADIOv2.5 is the closest to DINOv2 of the same size of any of the current agglomerative models. We train with a learning rate of $1e-3$, and use a weight of $1$ for all tasks. We purposefully don't tune any hyperparameters.

\begin{table}[h]
    \centering
    \begin{tabular}{r|ccccc}
    \hline
    Model & Backbone     & \shortstack{SemSeg\\mIoU} $\uparrow$ 
                                      & \shortstack{Depth\\RMSE} $\downarrow$ 
                                                    & \shortstack{Normal\\mErr} $\downarrow$ 
                                                                 & \shortstack{Boundary\\Loss} $\downarrow$  \\
    \hline
    MLoRE     & Custom   & 55.96      & 0.5076      & 18.33      & -           \\
    \hline
    DINOv2    & ViT-B/16 & \bf{60.64} & \bf{0.4816} & \bf{18.19} & 0.1268      \\
    Theia     & ViT-B/16 & 38.90      & 0.6377      & 24.11      & 0.1298      \\
    UNIC      & ViT-B/16 & 42.21      & 0.6172      & 22.78      & 0.1285      \\
    RADIOv2.5 & ViT-B/16 & 57.19      & 0.4980      & 20.04      & \bf{0.1263} \\
    \hline
    DINOv2    & ViT-L/14 & \bf{62.94} & \bf{0.4406} & \bf{17.63} & 0.1266      \\
    UNIC      & ViT-L/14 & 58.56      & 0.4916      & 19.34      & 0.1274      \\
    RADIOv2.5 & ViT-L/16 & 61.42      & 0.4577      & 18.57      & \bf{0.1259} \\
    \hline
    AM-RADIO  & ViT-H/16 & 62.76      & \bf{0.4339} & 18.43      & 0.1266      \\
    RADIOv2.5 & ViT-H/16 & \bf{63.82} & 0.4353      & \bf{17.67} & \bf{0.1256} \\
    \hline
    DINOv2    & ViT-g/14 & 63.89      & 0.4252      & 17.20     & 0.1262       \\
    \hline
    \end{tabular}
    \caption{Multi-task dense results on the NYUDv2 dataset. Note that we report the ``Boundary Loss'' and not the osdF metric due to the latter having a dependency on Matlab to compute.}
    \label{tab:dense-tasks-nyud}
\end{table}

\begin{table}[h]
    \centering
    \begin{tabular}{r|cccccc}
    \hline
    Model & Backbone     & \shortstack{SemSeg\\mIoU} $\uparrow$ 
                                      & \shortstack{Parsing\\mIoU} $\uparrow$ 
                                                    & \shortstack{Saliency\\maxF} $\uparrow$ 
                                                                 & \shortstack{Normal\\mErr} $\downarrow$
                                                                               & \shortstack{Boundary\\Loss} $\downarrow$ \\
    \hline
    MLoRE     & Custom   & 81.41      & 70.52       & \bf{84.90} & \bf{13.51}  & -         \\
    \hline                   
    DINOv2    & ViT-B/16 & 81.68      & \bf{73.24}  & 77.54      & 17.44       & 0.0619      \\
    Theia     & ViT-B/16 & 69.84      & 60.67       & 80.63      & 16.94       & 0.0623      \\
    UNIC      & ViT-B/16 & 75.90      & 62.85       & \bf{81.84} & \bf{15.78}  & 0.0620      \\
    RADIOv2.5 & ViT-B/16 & \bf{81.75} & 71.49       & 81.26      & 16.10       & \bf{0.0618} \\
    \hline                   
    DINOv2    & ViT-L/14 & 81.97      & \bf{74.51}  & 77.22      & 17.77       & 0.0620      \\
    UNIC      & ViT-L/14 & 81.82      & 72.24       & 79.21      & 17.35       & 0.0621      \\
    RADIOv2.5 & ViT-L/16 & \bf{82.87} & 74.32       & 81.65      & 16.15       & \bf{0.0617} \\
    \hline                   
    AM-RADIO  & ViT-H/16 & 82.78      & 74.42       & 78.48      & 17.53       & 0.0619      \\
    RADIOv2.5 & ViT-H/16 & \bf{83.43} & \bf{75.75}  & 81.19      & 16.16       & \bf{0.0617} \\
    \hline                   
    DINOv2    & ViT-g/14 & 82.47      & 75.56       & 76.93      & 17.59       & 0.0619      \\
    \hline
    \end{tabular}
    \caption{Multi-task dense results on the PASCAL Context dataset. Note that we report the ``Boundary Loss'' and not the osdF metric due to the latter having a dependency on Matlab to compute. We also don't bold a cell in a model size group if a smaller model has even higher quality for a given benchmark.}
    \label{tab:dense-tasks-pascal}
\end{table}

\section{Hyperparameters}

\begin{table}[h]
    \centering     
    \begin{tabular}{lccc}    
    \toprule
        \textbf{Parameter}       & \multicolumn{2}{c}{\textbf{Value}} \\
             & RADIOv2.5-L & RADIOv2.5-H \\
        \midrule  
        Backbone & ViT-L & ViT-H \\
        Learning Rate   & \multicolumn{2}{c}{$1e^{-3}$}    \\        
        Weight Decay  & \multicolumn{2}{c}{$2e^{-2}$}    \\ 
        Teachers & \multicolumn{2}{c}{DFN CLIP} \\
         &  \multicolumn{2}{c}{SigLIP 400m} \\
         &  \multicolumn{2}{c}{DINOv2-g-reg} \\
         &  \multicolumn{2}{c}{SAM-H} \\
         Feature Normalization & \multicolumn{2}{c}{PHI-S} \\
         Dataset & \multicolumn{2}{c}{DataComp-1B} \\
         Feature Distillation Loss & \multicolumn{2}{c}{MSE} \\
         Summary Loss & \multicolumn{2}{c}{Cosine} \\
         Backbone pre-training & \multicolumn{2}{c}{ImageNet-1k} \\
         
    \bottomrule
    \end{tabular}  
    \caption{RADIOv2.5 Training Hyperparameters}
    \label{tab:hyperparameters}
\end{table}

In Table~\ref{tab:hyperparameters}, we report our RADIOv2.5 training parameters.

\section{A Measure of Scale Equivariance}\label{apdx:scale_eq}

We define a measure of scale equivariance $\sigma^2_{\text{scale}}$: given an array of feature tensors $\{F_i\}$ with shapes $(H_i, W_i, C)$:

1. Let $F_{\text{min}}$ denote the tensor with the smallest spatial dimensions $(H_{\text{min}}, W_{\text{min}}, C)$.

2. Compute the per-channel mean and variance of $F_{\text{min}}$:
   \begin{equation}
   \mu_c = \frac{1}{H_{\text{min}} W_{\text{min}}} \sum_{h=1}^{H_{\text{min}}} \sum_{w=1}^{W_{\text{min}}} F_{\text{min}}(h, w, c)
   \end{equation}
   \begin{equation}
   \sigma_c^2 = \frac{1}{H_{\text{min}} W_{\text{min}}} \sum_{h=1}^{H_{\text{min}}} \sum_{w=1}^{W_{\text{min}}} \left(F_{\text{min}}(h, w, c) - \mu_c \right)^2
   \end{equation}

3. Normalize each tensor $F_i$ using $\mu_c$ and $\sigma_c$:
   \begin{equation}
   \hat{F}_i(h, w, c) = \frac{F_i(h, w, c) - \mu_c}{\sigma_c}
   \end{equation}

4. Bilinearly interpolate each normalized tensor $\hat{F}_i$ down to $(H_{\text{min}}, W_{\text{min}}, C)$, resulting in tensors $\{\tilde{F}_i\}$.

5. Stack all resized tensors $\{\tilde{F}_i\}$ along a new dimension and compute variance along this new dimension:
   \begin{equation}
   \sigma^2(h, w, c) = \text{Var}(\{\tilde{F}_i(h, w, c)\})
   \end{equation}

6. Finally, compute the average variance over the spatial dimensions:
   \begin{equation}
   \sigma^2_{\text{scale}} = \frac{1}{H_{\text{min}} W_{\text{min}}} \sum_{h=1}^{H_{\text{min}}} \sum_{w=1}^{W_{\text{min}}} \sigma^2(h, w, c)
   \end{equation}

\subsection{Scale Variance Implementation}

\lstdefinestyle{pythonstyle}{
    language=Python,
    basicstyle=\ttfamily\small,    
    keywordstyle=\color{blue},     
    stringstyle=\color{orange},    
    commentstyle=\color{gray},     
    showstringspaces=false,        
    frame=single,                  
    breaklines=true,               
    numbers=left,                  
    numberstyle=\tiny\color{gray}  
}

\begin{lstlisting}[style=pythonstyle]

def scale_variance(tensors: List, scale_up: bool): 
    """Compute feature variance across scales.
    
    Steps:
    * Per-channel standardization using the stats (mean/std)
      of largest features (if scale_up) or smallest features (if scale_down)
    * Interpolation to the size of the largest features (if scale_up) or
      smallest features (if scale_down).
    * Stack along a new dimension.
    * Compute variance along the new dimension.
    * Average across batch and spatial dimensions.
    """
    
    if scale_up:
        target_tensor = max(tensors, key=lambda x: x.numel())
        # Find the largest spatial dimensions
        target_H = max(tensor.shape[1] for tensor in tensors)
        target_W = max(tensor.shape[2] for tensor in tensors)
    else:
        target_tensor = min(tensors, key=lambda x: x.numel())
        # Find the smallest spatial dimensions
        target_H = min(tensor.shape[1] for tensor in tensors)
        target_W = min(tensor.shape[2] for tensor in tensors)
        
    # Compute mean and std along spatial dimensions (H, W) for each channel
    mean = target_tensor.mean(dim=(1, 2), keepdim=True)
    std = target_tensor.std(dim=(1, 2), keepdim=True)

    # Normalize each tensor and resize to the largest spatial dimensions
    normalized_tensors = []
    for tensor in tensors:        
        # Mean-center and normalize
        normalized_tensor = (tensor - mean) / (std + 1e-8)  # Adding a small epsilon to avoid division by zero

        # Resize to (B, max_H, max_W, C)
        resized_tensor = F.interpolate(
            normalized_tensor.permute(0, 3, 1, 2), 
            size=(target_H, target_W), 
            mode='bilinear', 
            align_corners=False
        ).permute(0, 2, 3, 1)

        normalized_tensors.append(resized_tensor)    

    stacked_tensors = torch.stack(normalized_tensors)  # Shape: (num_tensors, B, max_H, max_W, C)

    # Compute variance along the first dimension (num_tensors)
    variance_tensor = torch.var(stacked_tensors, dim=0)
    
    return variance_tensor.mean().item()

\end{lstlisting}

\section{Mode-Switch PCA Visualizations}

\begin{figure}[h]  
    \centering
    \includegraphics[width=\textwidth]{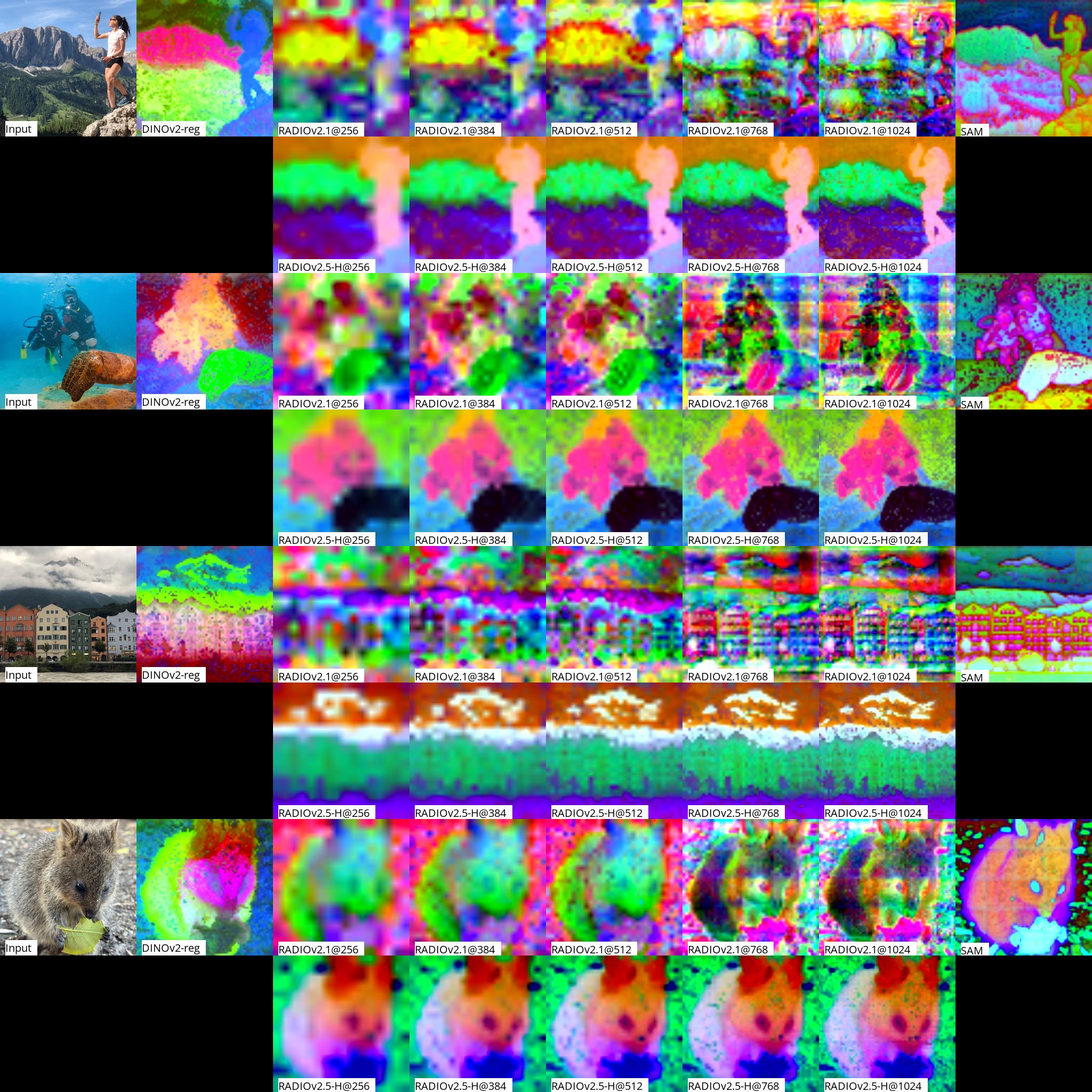}
    \caption{Visualizations of model features exhibiting the mode switch issue. We use PCA to project patch tokens into a 3D-space representing RGB colors. From left to right: input image, DINOv2, RADIO (baseline model) at 256x256, 384x384, 768x768, 1024x1024, and SAM. The visualizations illustrate how our baseline RADIO switches from producing DINO-like features at low resolution to producing SAM-like features at high resolution. }
    \label{fig:mode-switch-features-multiple}
\end{figure}

Figure \ref{fig:mode-switch-features-multiple} shows more visualization of the RADIO feature changes incurred by resolution increases.

\section{Visualizations of Native vs Emulated High-Resolution Inference}

\begin{figure}[h]  
    \centering
    \includegraphics[width=\textwidth]{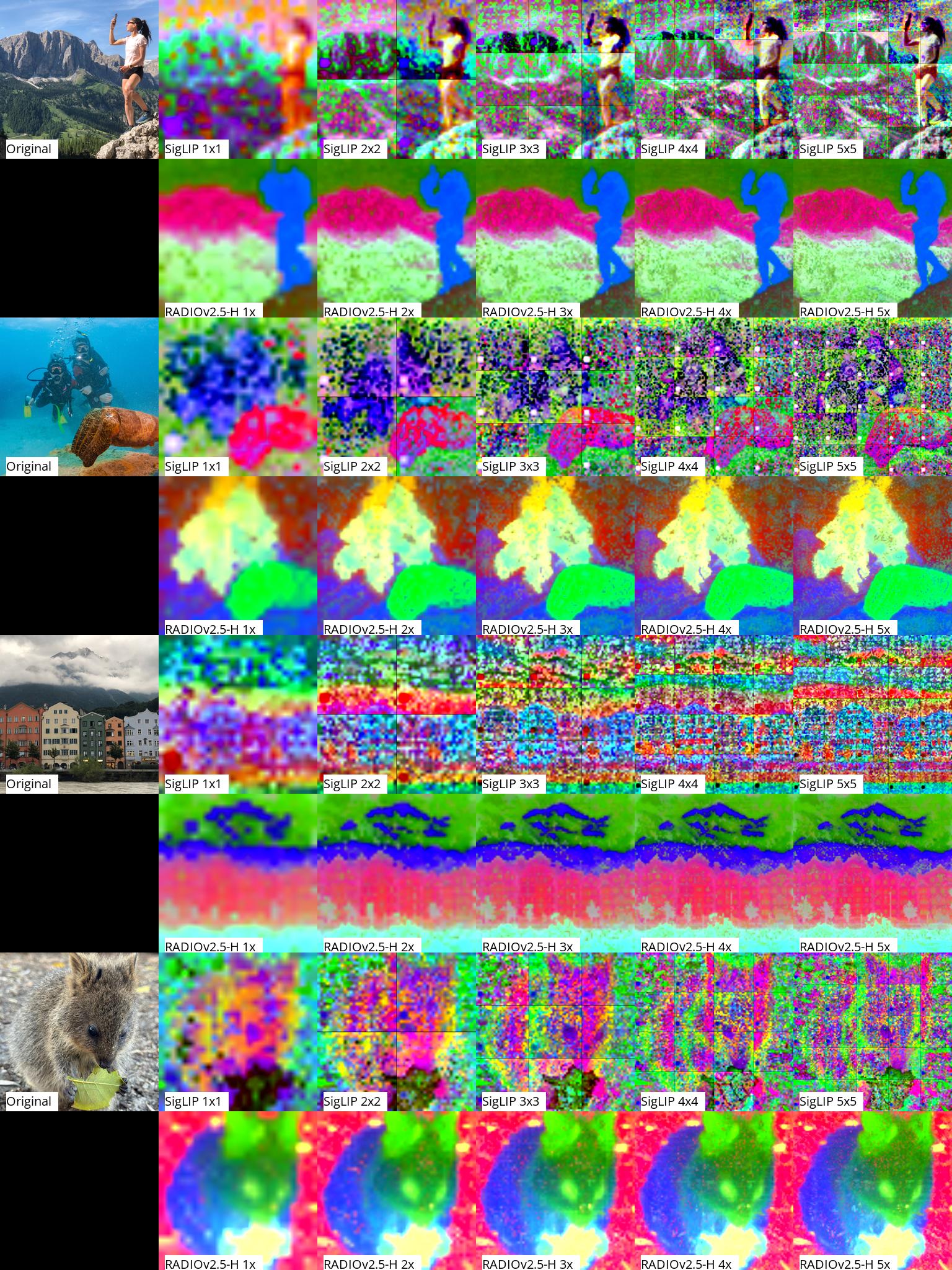}
    \caption{Visualization of image upscaling with tiling and SigLIP}
    \label{fig:tiling-multiple}
\end{figure}

Figure \ref{fig:tiling-multiple} shows visualizations of output features for emulated high-resolution inference through tiling, and for native high-resolution inference using RADIOv2.5. 


    

\section{Token Merging}

\subsection{Ablation Study on ToMe Parameters}

\begin{table}[]
    \begin{tabular}{lcccccc}
    
    \toprule
         \multirow{2}{*}{\textbf{Sink Layout}} &  \multicolumn{1}{|c|}{\B{SigLIP}} & \multicolumn{2}{c}{\B{RADIOv2.5-H}} \\
           &  \multicolumn{1}{|c|}{\textit{Values}} & \textit{Keys} & \textit{Values} \\
        \midrule
        $4 \times 4$ &   \multicolumn{1}{|c|}{0.56}  & 0.54 & \B{0.50} \\
        $6 \times 6$ &   \multicolumn{1}{|c|}{0.52} & 0.55 & \B{0.48}\\
         $8 \times 8$ &   \multicolumn{1}{|c|}{0.56} & 0.59 & \B{0.53} \\
                      
    \bottomrule
    \end{tabular}
        
    \caption{Reconstruction error (normalized MSE) after token merging/unmerging, using as criteria the "keys" (we use the attention keys) or "values" (we use the patch token values). The ``Sink Layout'' column indicates the strided arrangements for the merge sinks. RADIOv2.5-H exhibits a systematically lower error than SigLIP.}
    \label{tab:tome-reconstruction-error}
\end{table}

\subsection{More Token Merging Visualizations}

\begin{figure}[h]  
    \centering
    \includegraphics[width=\textwidth]{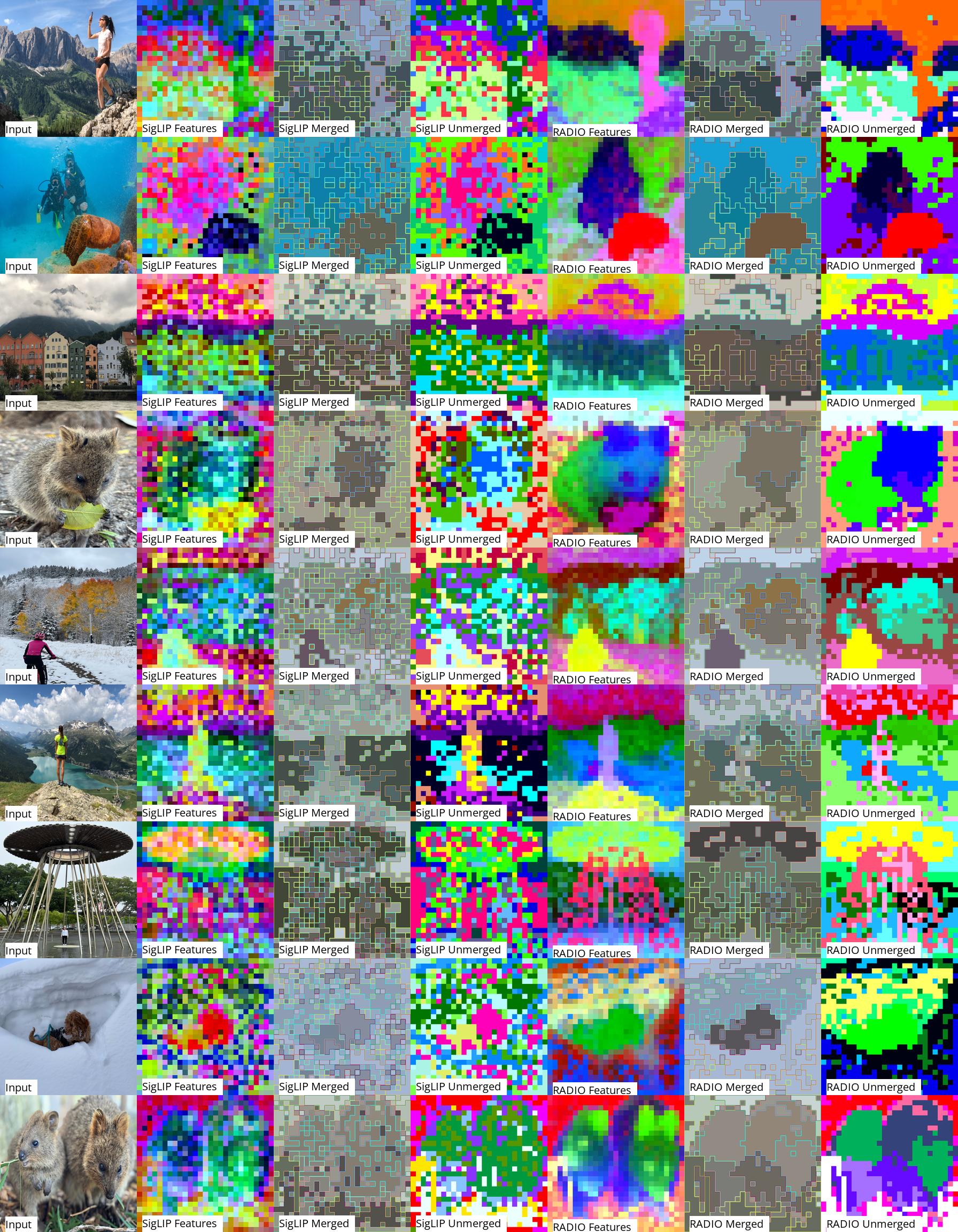}
    \caption{ToMe visualizations }
    \label{fig:tome-multiple}
\end{figure}

Figure \ref{fig:tome-multiple} shows more visualization of the ToME compression/decompression. Each input image yields 27x27=729 tokens, which are then compressed to 9 tokens using ToME.

\section{Mosaic Visualizations}

\begin{figure}[h]  
    \centering
    \includegraphics[width=\textwidth]{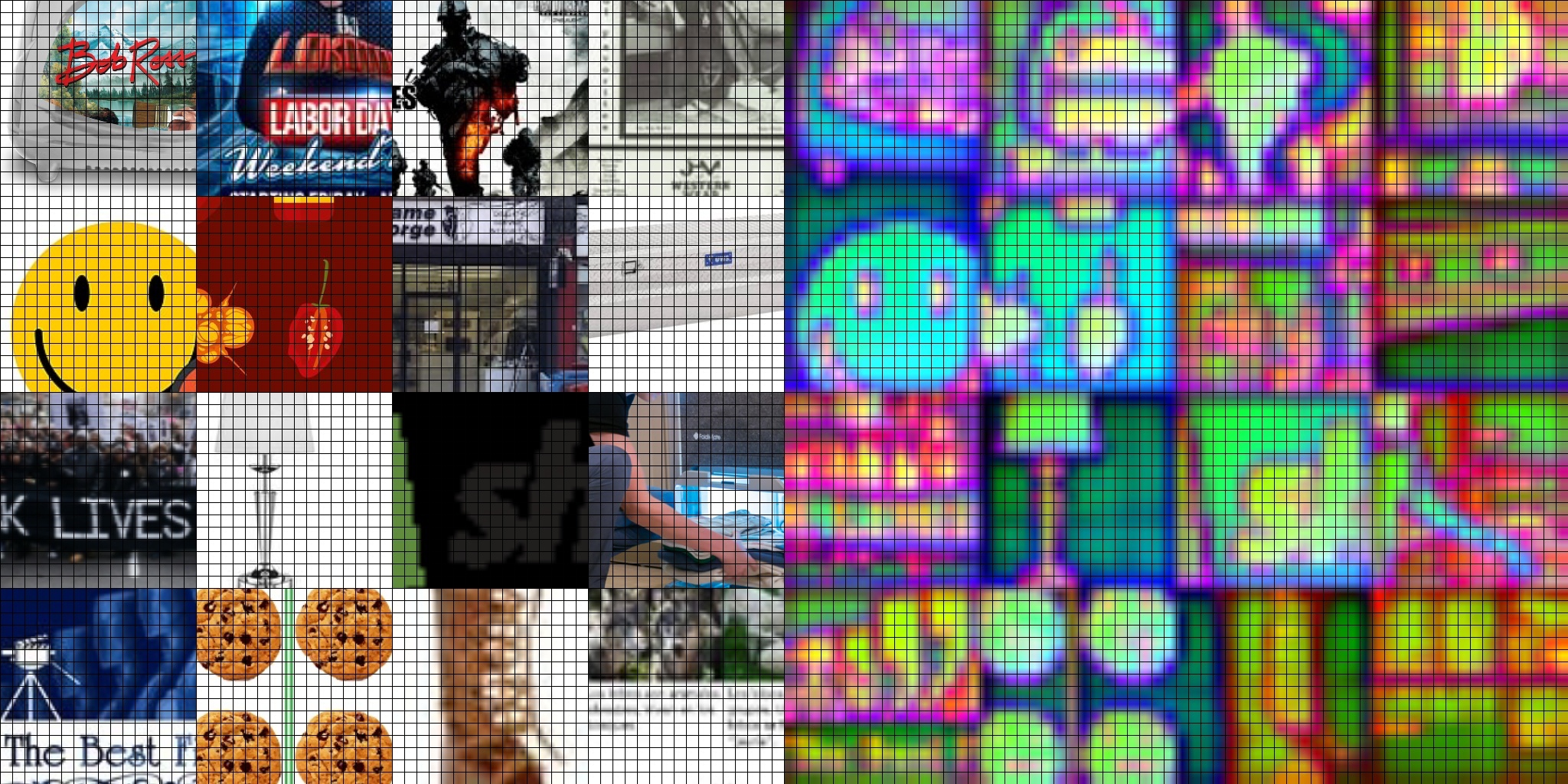}
    \caption{Mosaic visualization in a $4 \times 4$ arrangement }
    \label{fig:mosaic-and-features-4x4}
\end{figure}

\begin{figure}[h]  
    \centering
    \includegraphics[width=\textwidth]{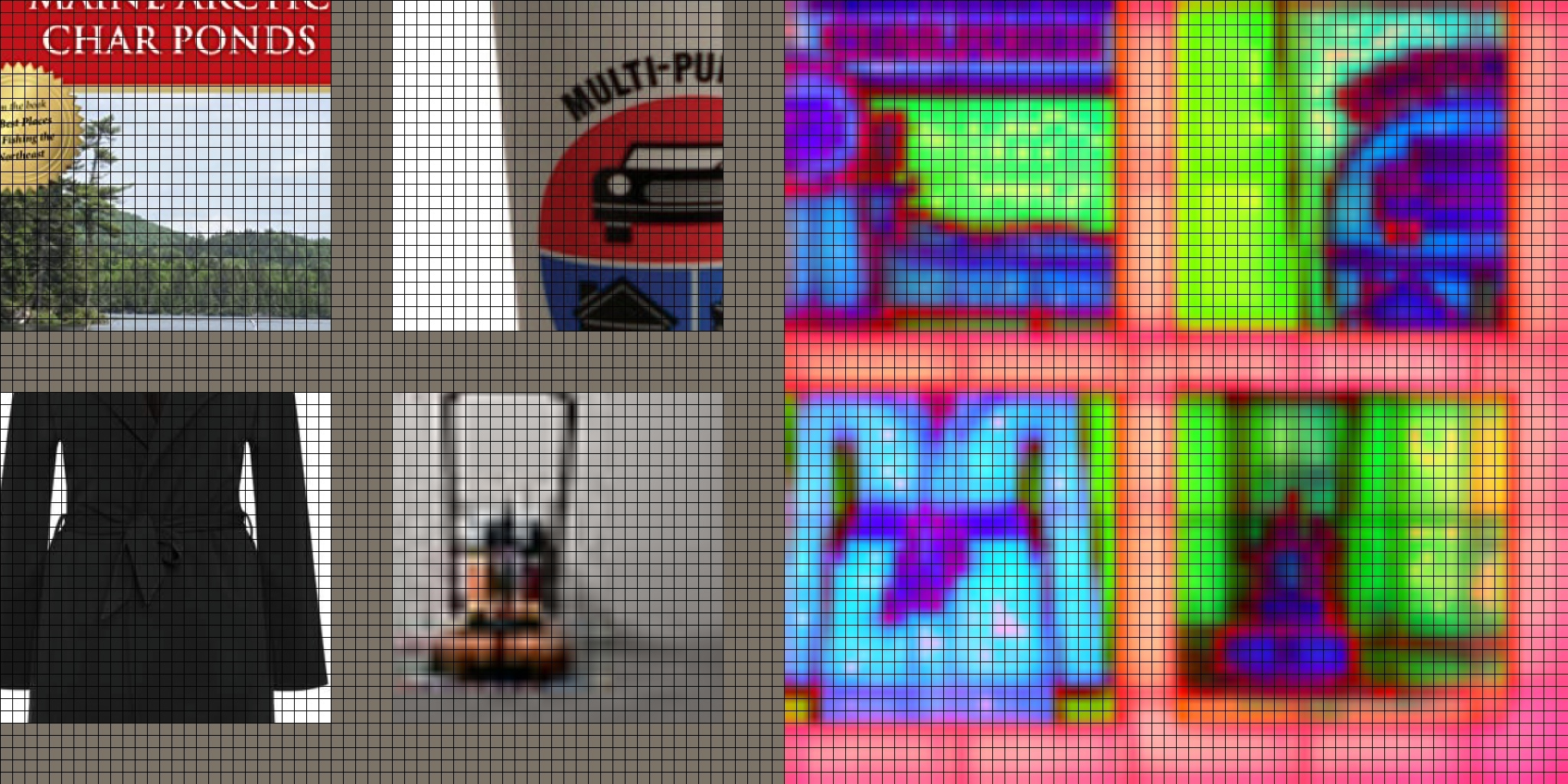}
    \caption{Mosaic visualization in a $2 \times 2$ arrangement }
    \label{fig:mosaic-and-features-2x2}
\end{figure}

Figures \ref{fig:mosaic-and-features-4x4} and \ref{fig:mosaic-and-features-2x2} show visualizations of mosaic augmentations under $2 \times 2$ and $4 \times 4$ arrangements.

\end{document}